\documentclass{article}
\usepackage{iclr2027_conference,times}

\usepackage[T1]{fontenc}
\usepackage{microtype}
\usepackage{graphicx}
\usepackage{booktabs}
\usepackage{multirow}
\usepackage{array}
\usepackage{tabularx}
\usepackage{placeins}
\usepackage{amsmath,amssymb,amsfonts}
\usepackage[table]{xcolor}
\usepackage{xspace}
\usepackage{hyperref}
\usepackage{url}
\hypersetup{colorlinks=true,allcolors=black}

\newcommand{\datasetname}{\textsc{ChronoDossier}\xspace}
\newcommand{\docmode}{\textsc{Full}\xspace}
\newcommand{\ragmode}{\textsc{Retrieve}\xspace}
\newcolumntype{Y}{>{\raggedright\arraybackslash}X}
\definecolor{bestshade}{HTML}{DCEFE5}
\newcommand{\bestscore}[1]{\cellcolor{bestshade}\textbf{#1}}

\usepackage{etoolbox}
\AtBeginEnvironment{table}{\setlength{\abovecaptionskip}{0pt}\setlength{\belowcaptionskip}{11pt}}

\title{\datasetname: Long-Form Personas as\\Temporal Memory Representations}

\author{Noah Sturgill \quad Jinho D. Choi \\
Emory University}

\iclrfinalcopy

\begin{document}
\maketitle
\lhead{Preprint. Under review.}

\begin{abstract}
Long-horizon agents rely on stored records to recover a person's state as facts change. Reliable answers depend on how facts are recorded, retrieved, and presented to the reader. We treat a long-form persona dossier as a memory representation of a known latent state, an invariant identity plus a timeline of facts, routines, and decision rules tied to life phases. We present \datasetname, a benchmark of 311 synthetic dossiers of 1k--10k words written by seven generators from dated timelines. Intrinsic metrics evaluate the records against their intended states; extrinsic metrics test recovery by five readers on a seeded 32-persona GPT-5.1 panel balanced across lengths (520 rated turns per condition). The linked states, records, evidence packets, and answers support tests of construction and evidence delivery. In a contextual-stress control, tag-scoped writer inputs remove detected out-of-scope anchor copying. Frozen questions authored on staged dossiers transfer less reliably to one-pass dossiers of the same states. In matched evidence controls, whole sections repair 29 of 32 GPT-5.1 and 25 of 30 Sonnet 4.6 failures labeled as packaging, compared with 2 of 8 and 1 of 8 answer-formation failures. Attribution uses separate key-based and key-free support measures, with model disagreements retained. Length remains confounded with persona and question composition; the observed between-bucket widening depends on excerpt construction. The benchmark supports offline evaluation against known facts. Memory acquisition from interaction and validation against real users remain outside its scope.
\end{abstract}

\section{Introduction}
\label{sec:introduction}

Agents that serve one person over time answer, plan, and advise from stored records about that person, such as profiles and summaries \citep{park2023generative,packer2023memgpt}. In agentic use, a memory error changes what the agent does: answering from a superseded role, an outdated routine, or an unsupported premise acts on the wrong state. Reliability requires preserving intended distinctions when a record is written and recovering relevant evidence when a later decision needs it. Time complicates both stages: a fact can hold in one life phase and not the next. Most existing benchmarks measure recovery. LongMemEval tests extraction, multi-session reasoning, temporal reasoning, updates, and abstention \citep{wu2025longmemeval}; AMemGym evaluates interactive state evolution \citep{cheng2026amemgym}; and long-context suites show that nominal context capacity does not ensure task performance \citep{yen2025helmet}. In these settings, the record usually arrives fully formed. Synthetic-persona generation exposes the preceding problem: diverse seeds can yield realizations that converge, leak facts between sections, or blur which phase governs a claim.

We study a long-form persona dossier as a memory representation. Each dossier encodes a structured latent state, invariant identity plus a dated timeline of facts, routines, and decision rules, and that state is known because the dossier is generated from it. \emph{Intrinsic} metrics compare the representation with its source state and with other personas' representations; \emph{extrinsic} metrics ask whether a model reading the representation recovers the state across dossier lengths, as access narrows to retrieved passages, and as the governing life phase changes. The study asks which intended distinctions the constructed representation encodes and whether it keeps personas and life phases apart; how well models recover the encoded state as access and evidence delivery change; and whether interventions in the encoding, evidence, or model's weights improve recovery. The current design leaves an independent length effect unresolved because personas and questions differ across length buckets.

We present \datasetname, a benchmark with three parts. It \emph{constructs} the representation, turning a structured temporal seed into a provenance-preserving artifact graph whose views encode the same state. A \emph{metric suite} (Section~\ref{sec:metrics}) evaluates that representation intrinsically and extrinsically; gold-section labels and answer keys attribute explicitly supported failures to retrieval, packaging, or answer formation. \emph{Replay} tests how well five models carry the representation into answers from the full dossier (\docmode) or retrieved passages (\ragmode), under alternative encodings of the same evidence, and after parameter-efficient adaptation, which tests whether a model can learn to use it.

The controls connect construction and evidence delivery to recovery. At 10k words, frontier \ragmode accuracy in the main configuration is 0.705 to 0.745 of \docmode accuracy, a between-bucket association whose widening shrinks when sections are delivered whole. The paired comparison shares personas, items, and scripted histories; earlier live answers differ by condition. The packaging diagnosis predicts which failures whole sections repair. Construction choices also change source-item transfer between records of the same states. The losses concentrate in particular question types: artifact-and-workflow questions are hardest, compound questions lose most under retrieval, and explicit facts re-asked after scripted history account for most of the retrieval loss. For an open-source reader, parameter-efficient adaptation lowers inference accuracy, while overall and explicitly supported accuracy have no detected change. The claims concern this task and artifact format. Replay dossiers and items are written by GPT-5.1, with an exploratory 16-persona panel written by Sonnet 4.6; memory acquisition from real interactions, human realism, psychological validity, and deployment utility remain outside the study.

\FloatBarrier
\section{Related Work and Positioning}

\textbf{Persona resources and behavioral evaluation.} PERSONA-CHAT established profile-conditioned dialogue \citep{zhang2018persona}, and Dialogue NLI and later personalization work made persona consistency and implicit inference measurable \citep{welleck2019dialoguenli,madotto2019personalizing,zhou2023predictpersona}. Persona Hub scales persona data; RoleLLM/RoleBench and OpenCharacter support role-play \citep{ge2024personahub,wang2024rolellm,wang2025opencharacter}. Among behavioral evaluations, InCharacter uses psychological interviews \citep{wang2024incharacter}; PersonaGym and CharacterBench use persona-relevant environments and targeted questions \citep{samuel2024personagym,zhou2025characterbench}; TimeChara tests knowledge at a selected narrative point \citep{ahn2024timechara}; and PersonaArena, SPASM, and PersonaForge study context-rich or long-horizon behavior \citep{shi2026personaarena,luo2026spasm,tong2026personaforge}. \datasetname examines the representation upstream of these behaviors: a long-form dossier with inspectable provenance and claims that change across explicit life phases. Replay accuracy measures artifact-internal recoverability and does not measure human simulation fidelity; Appendix Table~\ref{tab:positioning} compares representation scale across these resources.

\textbf{Long-context, memory, and temporal evaluation.} Long-context benchmarks test retrieval, reasoning, and robustness as evidence grows. LongMemEval decomposes long-term conversational memory into information extraction, multi-session reasoning, temporal reasoning, updates, and abstention \citep{wu2025longmemeval}; LoCoMo evaluates question answering and event summarization over long conversations \citep{maharana2024locomo}; and AMemGym adds structured state evolution and on-policy interaction \citep{cheng2026amemgym}. MSC extends persona dialogue across sessions \citep{xu2022goldfish}, and PersonaMem and PrefEval test whether models track evolving user profiles and preferences \citep{jiang2025personamem,zhao2025prefeval}. Agent memory systems decide what to store and retrieve \citep{park2023generative,packer2023memgpt}, and the granularity of memory or retrieval units shapes what is found \citep{pan2025secom,chen2024densex}. TimeChara and temporal RAG benchmarks isolate point-in-time fidelity and retrieval over indexed evidence \citep{ahn2024timechara,chen2025chronoqa,cao2026re3}. Their inputs evolve as dialogue, news, or event streams. \datasetname fixes the record in advance: replay selects among co-existing life phases within one dossier, which isolates representation realization from the accumulation and retrieval of later history.

\textbf{Synthetic-data audits and documentation.} Synthetic-data quality is often summarized by similarity, diversity, or model-based judgments. Sentence embeddings \citep{reimers2019sentencebert} can expose semantic concentration that lexical overlap misses, and recent work evaluates synthetic data through downstream effects alongside surface quality \citep{fan2026optimsyn}. Data statements, datasheets, and model cards motivate release procedures that keep realized outputs inspectable \citep{bender2018datastatements,gebru2021datasheets,mitchell2019modelcards}. \datasetname applies both views to one representation: intrinsic audits and downstream replay share the same records.

\section{Persona Dossiers as Memory Representations}
\label{sec:construction}

\subsection{Latent state and its encodings}

A persona's latent state $s$ has invariant identity fields and a timeline of phase-bounded facts, routines, constraints, and decision rules. Seeds vary along 11 audited axes: age range, gender, ethnicity/background, region, urban form, climate context, education, wealth, job, specialization, and work schedule. Appendix Table~\ref{tab:seed-axes} lists their values and realized distribution; regions are setting descriptors, many US-oriented, and do not define a set of countries. This operational target does not claim demographic completeness.

\begin{figure}[h]
  \centering
  \includegraphics[trim=20 176 24 198,clip,width=0.91\textwidth]{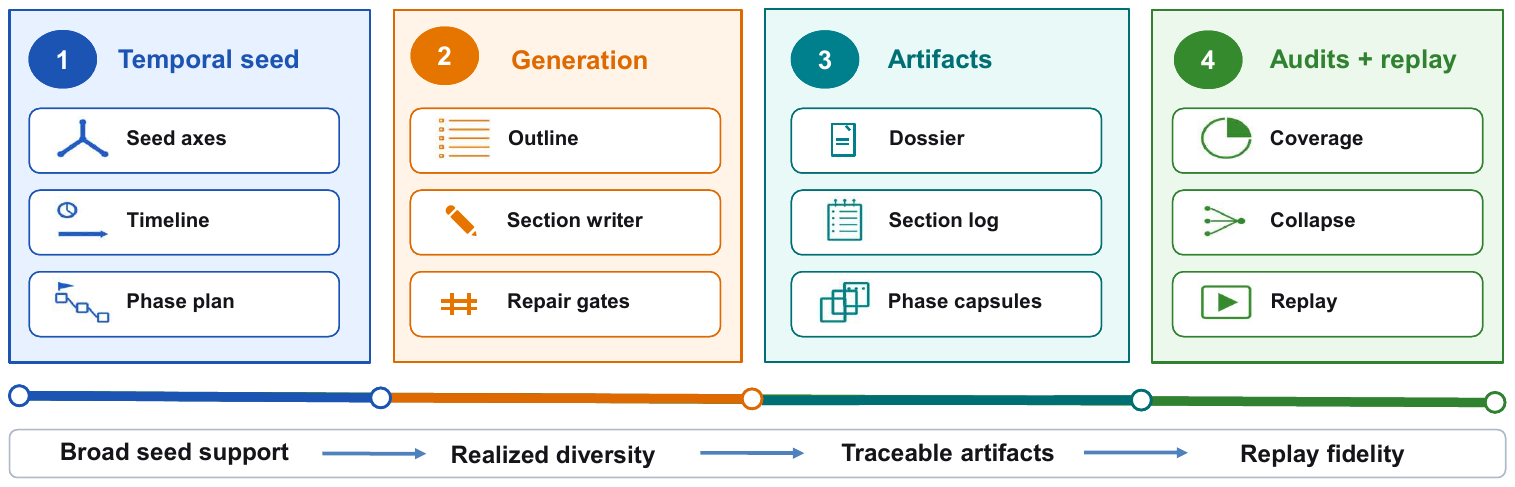}
  \caption{\datasetname joins construction and downstream use. A temporal seed is sliced into phase-local payloads; accepted sections become dossiers, cards, and phase capsules; audits and replay share their identifiers. \docmode supplies the full dossier and \ragmode three section passages.}
  \label{fig:framework}
\end{figure}

Construction encodes $s$ as a representation exported in five linked views (Figure~\ref{fig:framework}): (1) the long dossier used by \docmode; (2) a compact card of facts, values, heuristics, boundaries, and timelines; (3) section records with tags and time scope; (4) phase capsules sliced from multi-phase sections; and (5) repair logs. Every passage keeps its persona, section, and phase identifiers; a replay miss therefore links back to the frozen item, retrieved sections, card, capsule, and any pre-export repairs. Here, ``graph'' refers to this joinable provenance structure; no graph representation is learned. Retrieved evidence can be re-encoded in two further forms. \emph{Versioned state records} list each changing fact's current and superseded values under one key; a \emph{resolved-state packet} selects the same records but states the value that holds at the queried year and keeps only the immediately superseded version.

\textbf{Worked record.} One released 10k-word realization shows how the views encode one latent state and how a replay miss traces back through them. The seed places Daichi Lee in a mid-size riverfront suburb with a riverfront trail network and four distinct seasons, and his card states that he ``works as a senior construction manager in a remote-first construction-tech startup.'' One explicitly supported item asks for the suburb, the trail network, and the four seasons. The \docmode answer passes. One \ragmode turn retrieves sections \texttt{s11}, \texttt{s1}, and \texttt{s3} but omits the trail and climate details; another retrieves \texttt{s11}, \texttt{s6}, and \texttt{s1}, includes all required facts, and passes. The chain separates topical retrieval from supplying every required fact.

\subsection{Construction procedure and controls}

Before prose generation, an outline planner assigns sections to phase IDs, year bounds, and required topic tags. The section writer receives invariant identity and only the phase-local seed fields relevant to the section, plus a compact continuity state from accepted sections. Structural and temporal gates test tags, chronology, phase scope, and length; novelty gates compare a draft with accepted text. Failed sections are repaired or regenerated. Two matched controls isolate construction choices. The \emph{payload control} gives matched section requests either the full seed or only the section's own tag-scoped fields. The \emph{flat control} replaces timeline-first stage-local generation with one-pass generation while holding seeds, timelines, nominal lengths, frozen replay items, target model (GPT-5.1), and judge fixed.

\textbf{Data provenance and human involvement.} All records are synthetic. Seven generators write them from analyst-defined seeds: GPT-5.1, GPT-5 mini, Claude Sonnet 4.6, Claude Haiku 4.5, Gemma 4 26B-A4B \citep{gemmateam2026gemma4}, Llama 3.1 70B Instruct \citep{grattafiori2024llama3}, and Qwen3 32B \citep{qwen2025report} (Appendix Table~\ref{tab:config}). Short names omit versions; GPT denotes GPT-5.1. The resource includes no study participants or collected real-person profiles. Outputs are validated, repaired, and exported without manual rewriting. Humans define seed axes, rubrics, and the protocol; they also provide the limited adjudication in Section~\ref{sec:controls} and the phase-rubric calibration in Appendix~\ref{app:calibration}. Requested, completed, frozen, and rated counts remain separate.

\FloatBarrier
\section{Evaluating a Memory Representation}
\label{sec:metrics}
\label{sec:benchmark}

\begin{table}[htbp]
\caption{Metric suite. Intrinsic metrics need no reader; extrinsic metrics require one.}
\label{tab:metrics}
\centering
\small
\setlength{\tabcolsep}{4pt}
\begin{tabularx}{\textwidth}{@{}llY@{}}
\toprule
Metric & Type & Definition \\
\midrule
Quality & Intrinsic & Consensus judge score on an eight-criterion document rubric \\
Collapse & Intrinsic & Mean nearest-neighbor cosine between dossier embeddings (Top1), its 95th percentile, and pairs above 0.8 \\
Scope leakage & Intrinsic & Sections with an exact multiword seed value from outside their scope \\
Coverage & Intrinsic & Share of permitted in-scope seed anchors that the section realizes \\
Recoverability & Extrinsic & Pass rate of a reader under an access condition, over all rated turns \\
Source-item transfer & Extrinsic & Recoverability on items authored from a different realization of the same state \\
Length retention & Extrinsic & \ragmode accuracy divided by \docmode accuracy within a length bucket \\
Temporal gap & Extrinsic & Phase-linked minus timeless accuracy on explicitly supported items \\
Failure attribution & Extrinsic & Scored SF misses by packet and labeled-source support \\
\bottomrule
\end{tabularx}
\end{table}

\subsection{Intrinsic metrics}

Table~\ref{tab:metrics} lists the metric suite. GPT-5.1 and Claude Sonnet 4.6 judges share one document rubric (Appendix Table~\ref{tab:rubric}) covering internal and temporal consistency, seed fidelity, specificity, readability, bias avoidance, retrieval readiness, and length control. Gemini 3 Pro Preview is added as a third judge when their scores differ by over 15 points, unless both are below 70. Model judges are measurement instruments; their outputs are not gold standards \citep{liu2023geval,zheng2023judging}. Collapse uses all-MiniLM-L6-v2 dossier embeddings: high Top1 means that distinct latent states produced similar representations, and the 95th percentile exposes clusters that the mean hides; exact lexical 5-gram overlap adds a surface view. Leakage counts exact multiword seed phrases (anchors) copied from an axis outside the requested scope. Phase packets, one per capsule, also receive an ordinal A--F temporal-distinctness tier, calibrated against human raters in Appendix~\ref{app:calibration}.

\subsection{Readers and extrinsic metrics}
\label{sec:readers}

Replay uses 32 held-out GPT-5.1 source personas, a seeded sample of eight per 1k, 2.5k, 5k, and 10k target-length bucket (Appendix~\ref{app:replay-config}), and five readers: GPT-5.1 and Sonnet (the \emph{frontier} targets) and the open-source Qwen, Llama, and Gemma. GPT-5.1 authors item definitions once from the seed, card, timeline, dossier, and section excerpts under a JSON schema; accepted items are frozen before target-model runs. Of 431 frozen definitions, 208 are scheduled as rated turns; retention ladders re-ask anchor items and time-jump episodes revisit earlier-phase items, which yields 520 rated turns per condition. Of 8,880 turns across 96 episodes, 8,360 are scripted history and 520 are rated live turns (Appendix Table~\ref{tab:history}); replay therefore measures answers after controlled histories and excludes fully model-generated conversation.

Items fall into three families. \emph{Surface-faithful} (SF) answers are explicitly supported; \emph{inferred-faithful} (IF) answers require inference from stored values, heuristics, or patterns; and \emph{not-in-persona} (NIP) requests require refusal because the premise is unsupported, fabricated, or privacy-invasive. Each item stores its phase, key or rubric, and gold section labels. SF items use answer keys with fallbacks to atomic NLI and to the judge; NIP items use refusal checks; IF items receive a Boolean judgment from the GPT-5.1 judge. The atomic NLI prompt was revised in September 2026: the frontier rows of Table~\ref{tab:replay} and other runs completed before the revision use the original prompt, and the open-source reruns, the Section~\ref{sec:item-family} contrasts, and the controls added during discussion use the revised one, which accepts some SF answers that fail the key. Comparisons across the two prompts score SF by the answer key alone (SF$_\text{key}$). A pass must satisfy the family rule and any phase constraint. Replay scores the eligible user-visible answer; an unterminated reasoning block or empty final is a nonanswer and fails. Judge payloads omit reader identity and access mode, although evidence can reveal the mode.

\docmode supplies the entire dossier. \ragmode supplies an \emph{evidence packet} of the top three section passages, each capped at 550 characters, from hybrid retrieval over all-MiniLM-L6-v2 embeddings and lexical, heading, tag, and year matches. A section longer than the cap is split into sentence-like chunks; its excerpt starts at the chunk with the greatest lexical overlap with the query and adds neighboring chunks while they fit. Every reported \ragmode run loads the embedder; the submitted open-source rows, which did not, are kept as a configuration record (Appendix~\ref{app:replay-config}). A \emph{query planner} builds the query from the current request and the two previous user turns: frontier targets write a structured query, and open-source targets use a fixed heuristic planner, adopted after none of 77 planning calls in a Qwen3 32B pilot returned a parseable query within 180 tokens. Frontier readers answer without extended reasoning (GPT-5.1 at minimal effort, Sonnet at temperature 0) within 120 output tokens; open-source readers use greedy decoding within 220. Frontier and open-source rows therefore differ in pipeline settings as well as in the reader. The item, reader, scoring rule, and history stay fixed across access modes. Neither condition isolates retrieval or serves as an oracle: retrieval-augmented generation couples evidence selection with answer formation \citep{lewis2020rag}, and long-context models use relevant evidence unevenly across positions and distractors \citep{liu2024lost,yen2025helmet}. Retrieval diagnostics record gold phase hit@3 and section hit@3 (a top-three passage comes from the item's phase or a labeled gold section) and phase purity. Failure attribution applies each frozen SF answer key to the labeled source sections and to the realized packet: a failure is assigned to retrieval when no labeled section is selected, to packaging when a labeled section is selected but the packet fails the key, and to answer formation when the packet passes the key.

\subsection{Interventions on evidence, encoding, and weights}
\label{sec:evidence-interventions}

The \emph{gold-section condition} supplies only the sections labeled for each item; 408 of the 431 definitions carry section labels, all resolving to stored sections. The \emph{fixed-total} policy distributes at most 1,650 characters across the selected passages, separating passage allocation from context growth. The depth and adaptation contrasts each score both arms in one run on the 32-persona panel and cache judge decisions for identical responses; the format contrast compares runs that share items and retrieved sections. The \emph{depth contrast} raises $k$ from 3 to 12 at the 1,650-character total for GPT-5.1, Sonnet, and Gemma with the heuristic planner. The \emph{format contrast} gives Gemma the same retrieved sections as raw text or as versioned state records; on the eight-persona screen, it also compares versioned records with the resolved-state packet at matched budgets. The \emph{adaptation contrast} compares untuned Gemma with a rank-16 QLoRA adapter \citep{hu2022lora,dettmers2023qlora} trained by supervised fine-tuning (SFT), under byte-identical prompts with versioned state records. The adapter was trained for one epoch on 62 personas disjoint from every evaluation panel, with passing GPT-5.1 full-dossier responses as targets and answer keys or refusals as fallbacks.

Every rated turn contributes to the headline rates, including nonanswers. Availability is reported separately from answer-conditional quality. Proportions use Wilson intervals; overall rates and paired contrasts of frozen turns use length-stratified, persona-clustered bootstrap intervals (Appendix~\ref{app:replay-uncertainty}).

\section{Results}
\label{sec:results}

\subsection{Intrinsic: what the representation encodes}

\textbf{Collapse and quality diverge.} Document quality and collapse capture different properties of the 311 completed dossiers (Table~\ref{tab:construction}). The highest-quality generator (Sonnet) is not the least collapsed (Haiku), and Qwen's low judged quality coincides with a cluster of similar dossiers: Qwen is the only generator with pairs above the 0.8 alert threshold. Its worst alert links two distinct dossiers whose prose begins with near-identical instructions to ``rewrite the Biography section''; repeated task framing drives the similarity, and the personas do not share the implicated fact. Qwen stays the most collapsed in both views under equal-size subsampling. Surface deduplication alone would miss such concentration: matched Wikipedia biographies are less concentrated in embedding space than the synthetic documents at comparable lexical 5-gram overlap (Appendix~\ref{app:construction-audits}). All generators export the same artifact families under timeline-first planning, stage-local writing, and live repair; the documents remain model-specific pipeline outputs and are not matched semantic realizations.

\begin{table}[htbp]
\caption{Intrinsic metrics by generator. C/R is completed/requested. Quality is the mean consensus score: the judge-pair mean, or the three-judge median when Gemini is added. Top1 is mean nearest-neighbor cosine, p95 its upper tail, and Alerts counts pairs above 0.8.}
\label{tab:construction}
\centering
\small
\begin{tabular}{lrrrrr}
\toprule
Generator & C/R & Quality$\uparrow$ & Top1$\downarrow$ & p95$\downarrow$ & Alerts$\downarrow$ \\
\midrule
GPT-5.1 & 94/100 & 84.69 & 0.5797 & 0.6723 & 0 \\
GPT-5 mini & 37/40 & 82.70 & 0.5794 & 0.6402 & 0 \\
Sonnet 4.6 & 35/40 & \bestscore{88.06} & 0.5365 & 0.6064 & 0 \\
Haiku 4.5 & 37/40 & 81.92 & \bestscore{0.5232} & \bestscore{0.6029} & 0 \\
Gemma 4 26B-A4B & 32/40 & 63.41 & 0.6274 & 0.6889 & 0 \\
Llama 3.1 70B & 37/40 & 53.46 & 0.5696 & 0.6593 & 0 \\
Qwen3 32B & 39/40 & 47.77 & 0.7773 & 0.8364 & 12 \\
\bottomrule
\end{tabular}
\end{table}

\textbf{Scope and transfer.} In a contextual-stress control, writers given the full seed copy an anchor outside the section's topic tags in 29 of 100 sections; with tag-scoped fields, this occurs in 0 of 100. In-scope coverage changes by $+0.016$ (95\% CI $[-0.019,+0.051]$). Construction also changes transfer: frozen items pass on 95.5\% of 202 turns against staged dossiers and 84.2\% against one-pass dossiers of the same states ($-11.4$ points from the unrounded counts, persona-bootstrap 95\% CI $[-23.1,-3.4]$). These items were authored against staged dossiers; the comparison measures source-item transfer for the construction package. The payload control isolates tag scoping and measures exact out-of-scope anchor copying.

\subsection{Extrinsic: recovering the encoded state}
\label{sec:extrinsic}

\begin{figure}[htbp]
  \centering
  \includegraphics[width=0.93\textwidth]{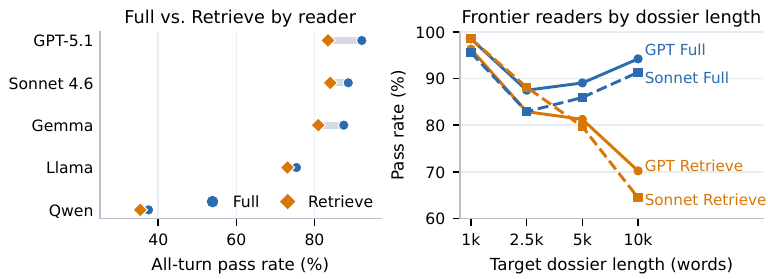}
  \caption{Left: all-turn replay pass rates after user-visible answer extraction, with nonanswers counted as failures. Right: frontier readers by dossier length. Retrieved-context accuracy is lower for longer dossiers although the top three passages include the correct phase for all 174 phase-labeled frontier items.}
  \label{fig:replay}
\end{figure}

Because Figure~\ref{fig:replay}'s left panel and Table~\ref{tab:replay} count nonanswers as failures, their access gaps also reflect interface completion; the right panel shows only frontier readers, which answer every turn.

\begin{table}[htbp]
\caption{Recoverability (\% pass) over 32 held-out personas, 520 rated turns per row, by score family. All-turn rates count nonanswers as failures. Open-source rows come from embedder-loaded reruns that score both access modes in one run (Appendix~\ref{app:replay-config}). SF$_\text{key}$ scores SF turns by the answer key alone; the frontier rows were judged with the original SF prompt and the open-source rows with the revised one. Appendix Tables~\ref{tab:replay-full} and~\ref{tab:replay-boot} give counts and CIs.}
\label{tab:replay}
\centering
\small
\begin{tabular}{llrrrrr}
\toprule
Access & Target & Overall & SF & IF & NIP & SF$_\text{key}$ \\
\midrule
\docmode & GPT & \bestscore{92.12} & 94.17 & 87.13 & 88.14 & 92.50 \\
\docmode & Sonnet & 88.65 & 91.94 & 89.11 & 67.80 & 90.00 \\
\ragmode & GPT & 83.46 & 80.83 & 91.09 & 86.44 & 83.89 \\
\ragmode & Sonnet & \bestscore{84.04} & 83.61 & 89.11 & 77.97 & 84.44 \\
\docmode & Qwen & 37.50 & 45.56 & 18.81 & 20.34 & 43.06 \\
\ragmode & Qwen & 35.38 & 40.00 & 30.69 & 15.25 & 38.06 \\
\docmode & Llama & 75.38 & 92.78 & 40.59 & 28.81 & 88.89 \\
\ragmode & Llama & 73.08 & 87.78 & 34.65 & 49.15 & 73.33 \\
\docmode & Gemma & 87.50 & 87.50 & 82.18 & 96.61 & 86.11 \\
\ragmode & Gemma & 80.96 & 80.56 & 73.27 & 96.61 & 74.72 \\
\bottomrule
\end{tabular}
\end{table}

\textbf{Length and excerpt construction.} Frontier \ragmode accuracy declines across the length buckets: it is 26 points lower for GPT and 34 for Sonnet on 10k-word than on 1k-word dossiers, against 4 under \docmode; at 10k words, \ragmode keeps 0.745 and 0.705 of \docmode accuracy (Appendix Table~\ref{tab:robustness}). The buckets contain different personas and questions. Full and Retrieve share scripted episodes, with earlier live answers generated separately. Across buckets, from 1k to 10k, the \ragmode-minus-\docmode gap changes by $-21.8$ points for GPT (length-stratified persona-bootstrap 95\% CI $[-40.8,-3.3]$) and $-29.9$ for Sonnet ($[-47.7,-12.0]$). The widening holds after reweighting toward a common observed item-family and subcategory mix, within SF identity-and-context items, under the Sonnet judge, after regenerating answers flagged as potentially truncated, under SF$_\text{key}$, and without SF items whose key the dossier does not clearly state (Appendix Table~\ref{tab:length-interaction}). Its size depends on evidence packaging. In a matched control, every variant retrieves the same three sections with the heuristic planner and a 220-token limit; under SF$_\text{key}$, the SF gap changes by $-24.7$ (GPT) and $-19.0$ points (Sonnet) with 550-character excerpts and by $-6.7$ and $-7.0$ with whole sections, whose intervals include zero (Appendix Table~\ref{tab:packaging}). Section selection and packet support measure different properties: the top three passages contain the item's phase for all 174 phase-labeled frontier items and a labeled section for 178 (GPT) and 183 (Sonnet) of the 188 section-labeled items (Appendix~\ref{app:supp-replay}). Labeled sections grow with dossier length (median 579 to 3,929 characters), and the share of a selected labeled section shown in the excerpt falls from 0.83 to 0.08 (GPT; 0.85 to 0.08 for Sonnet). Adding this share to a persona-clustered model of \ragmode SF accuracy shrinks the log-length coefficient from $-0.95$ to $-0.27$ (GPT) and from $-1.49$ to $-0.58$ (Sonnet), although share and length are nearly collinear in this design (Appendix Table~\ref{tab:coverage}). These results establish sensitivity to excerpt construction. The role of length remains unresolved; a larger comparison of the same states and questions across lengths would test it.

\textbf{Temporal gap.} In a post-hoc comparison of different item groups, under \ragmode, phase-linked SF items trail timeless ones by 12.7 points for GPT and 11.7 for Sonnet, against small, mixed-sign \docmode gaps (Appendix Table~\ref{tab:robustness}). This split leaves question difficulty and temporal demands confounded; the small IF and NIP cells limit comparisons across families.

\textbf{Open-source readers.} For open-source readers, returning an answer is itself a failure source. Qwen exposes an eligible answer on only 272 of 520 \docmode and 284 of 520 \ragmode turns, often stopping inside an unterminated reasoning block, and nonanswers account for most of its failures; on the SF turns it answers under \docmode, it passes 88.6\%, against Llama's 92.8\% (Appendix Table~\ref{tab:availability}). Llama and Gemma answer every turn; removing text after Gemma's end-of-turn marker changes its rates by $+0.8$ (\docmode) and $+2.1$ points (\ragmode) and leaves Llama's unchanged. Gemma leads open-source overall and answer-conditional performance; Llama leads open-source SF but passes under half of IF and NIP turns. Overall open-source \ragmode losses are 2.1 to 6.5 points, and every persona-bootstrap interval includes zero (Appendix Table~\ref{tab:replay-boot}). Under SF$_\text{key}$, \ragmode lowers SF accuracy by 15.6 points for Llama ($[-25.8,-6.3]$) and 11.4 for Gemma ($[-22.7,-0.3]$); Qwen's 5.0-point loss has an interval that includes zero ($[-14.6,+4.7]$; Appendix Table~\ref{tab:os-key}). Gemma is therefore the only open-source reader with consistent performance across item families, and the interventions in Section~\ref{sec:item-family} use it. Answer formation, interface completion, and memory remain separate failure sources.

\subsection{Which questions resist recovery}
\label{sec:qtypes}

Table~\ref{tab:qtypes} breaks the 520 core turns down by question type; three types stand out.

\begin{table}[htbp]
\caption{Pass rates (\%) by question type on the 520 core turns. Subcategories with fewer than 15 turns are omitted. All three readers use the runs of Table~\ref{tab:replay}; Gemma is scored with its extraction rule. Bold: paired exact McNemar $p<0.05$ for \docmode against \ragmode.}
\label{tab:qtypes}
\centering
\small
\setlength{\tabcolsep}{3.5pt}
\begin{tabular}{lrrrrrrr}
\toprule
& & \multicolumn{2}{c}{GPT} & \multicolumn{2}{c}{Sonnet} & \multicolumn{2}{c}{Gemma} \\
\cmidrule(lr){3-4}\cmidrule(lr){5-6}\cmidrule(lr){7-8}
Question type & $n$ & \docmode & \ragmode & \docmode & \ragmode & \docmode & \ragmode \\
\midrule
SF: identity and context & 266 & 97.4 & \textbf{84.6} & 97.7 & \textbf{86.5} & 92.1 & \textbf{85.7} \\
SF: phase-specific history & 46 & 97.8 & \textbf{82.6} & 91.3 & 89.1 & 89.1 & 78.3 \\
SF: artifacts and workflows & 34 & 79.4 & \textbf{58.8} & 67.6 & 55.9 & 70.6 & \textbf{50.0} \\
IF: decision heuristics & 65 & 86.2 & 87.7 & 86.2 & 92.3 & 81.5 & 69.2 \\
IF: values and goal trade-offs & 15 & 93.3 & 100.0 & 93.3 & 93.3 & 93.3 & 100.0 \\
NIP: false timeline or role claim & 17 & 82.4 & 64.7 & 58.8 & 70.6 & 88.2 & 88.2 \\
NIP: prompt injection & 33 & 97.0 & 97.0 & 69.7 & 78.8 & 100.0 & 100.0 \\
\midrule
Atomic questions & 395 & 93.9 & \textbf{86.6} & 90.4 & 86.8 & 90.6 & \textbf{85.3} \\
Compound questions & 125 & 86.4 & \textbf{73.6} & 83.2 & 75.2 & 77.6 & \textbf{67.2} \\
\midrule
Retention episodes (all SF) & 256 & 93.4 & \textbf{77.3} & 92.2 & \textbf{80.9} & 85.5 & 80.1 \\
Time-jump episodes & 136 & 89.7 & 88.2 & 88.2 & 86.8 & 84.6 & 79.4 \\
Adversarial episodes & 128 & 92.2 & 90.6 & 82.0 & 87.5 & 94.5 & \textbf{84.4} \\
\bottomrule
\end{tabular}
\end{table}

\textbf{Artifacts and workflows are hardest.} Among explicitly supported items, questions about the artifacts and workflows a persona relies on (for example, which planning or tracking templates a persona used as of a given year) have the lowest pass rates for every reader under both access modes. For the frontier readers, they are the only tabulated SF type below 90\% under \docmode.

\textbf{Compound questions lose most under retrieval.} Two-fact questions lose 8.0 to 12.8 points under \ragmode, against 3.6 to 7.3 for atomic ones, consistent with fragment-integration failure.

\textbf{Retrieval loss sits in re-asked facts.} Most of the frontier SF loss under retrieval falls on facts that retention episodes re-ask after scripted history: these episodes hold 47 of GPT's 55 and 37 of Sonnet's 43 SF turns that pass only under \docmode, while time-jump episodes change by two turns per frontier reader. Even identity and context items, the largest SF type, fall 6 to 13 points under \ragmode for all three readers from high \docmode accuracy. False-premise handling varies by reader: even under \docmode, Sonnet passes 58.8\% of false timeline or role claims and 69.7\% of prompt injections. Appendix~\ref{app:supp-replay} describes three recurring failure patterns.

\subsection{Changing the encoding, the evidence, or the weights}
\label{sec:item-family}
\label{sec:formats}

All Gemma contrasts in this section use the extraction rule of Table~\ref{tab:replay} (Appendix~\ref{app:interventions}). On the 32-persona panel, no packet encoding reaches full-dossier accuracy, and only versioned state records change overall accuracy under the exact test. Only within-row changes are interpreted.

\textbf{Encoding.} Re-encoding Gemma's retrieved sections as versioned state records lowers accuracy by 4.0 points (exact McNemar $p=0.038$; persona-bootstrap CI $[-9.5,+1.7]$) and NIP accuracy by 13.6 ($p=0.008$); every packet encoding stays 5.2 to 10.6 points below the full dossier (Appendix Table~\ref{tab:formats}). On the eight-persona screen, a resolved-state packet, which states the value that holds at the queried year, does not differ significantly from versioned records (120 against 115 of 154 turns at 8,000 characters, $p=0.38$).

\textbf{Depth.} Increasing $k$ from 3 to 12 changes the evidence mix even though the total packet length is fixed: gold-section recall rises from 0.64 to 0.97, and the amount of gold-section text per turn falls by half. Gemma's IF accuracy changes by $+5.9$ points ($p=0.24$; CI $[0.0,+11.7]$), and its SF and overall accuracy do not change. GPT and Sonnet already pass 92 and 97 of the 101 IF turns, respectively, and neither changes significantly (Table~\ref{tab:item-family}). Their gains are confined to cases where top-3 retrieval found no gold section (GPT $+9/-0$, $p=0.004$).

\textbf{Weights.} The SFT adapter lowers IF accuracy by 14.9 points ($p=0.011$, CI $[-26.5,-3.7]$); NIP accuracy rises by 11.9 ($p=0.065$; CI $[+1.6,+22.2]$), and SF and overall accuracy do not change significantly. Lost IF answers are mostly incomplete: adapted IF answers average 42.5 words against 88.7 untuned, and the judge cites omissions for 18 of the 23. On the eight-persona screen, none of ten adapters from five recipes differs significantly from the untuned model, overall (102 to 110 of 154 turns, against 108 and 106 for two gradings) or on IF (17 to 23, against 22).

\begin{table}[htbp]
\caption{Paired pass counts on the 520 rated turns of Table~\ref{tab:replay}; headers give turn counts. Depth rows fix total evidence at 1,650 characters and use the heuristic planner; adaptation uses versioned state records in both arms. Each row is one run with cached judge calls; Gemma rows use the extraction rule of Table~\ref{tab:replay}. Bold: exact McNemar $p<0.05$.}
\label{tab:item-family}
\centering
\small
\setlength{\tabcolsep}{4pt}
\begin{tabular}{lrrrr}
\toprule
Contrast & IF (101) & SF (360) & NIP (59) & Overall (520) \\
\midrule
GPT-5.1, $k$ 3$\to$12 & 92$\to$96 & 315$\to$312 & 54$\to$48 & 461$\to$456 \\
Sonnet 4.6, $k$ 3$\to$12 & 97$\to$94 & 314$\to$302 & 43$\to$46 & 454$\to$442 \\
Gemma, $k$ 3$\to$12 & 83$\to$89 & 288$\to$282 & 57$\to$59 & 428$\to$430 \\
Gemma, untuned$\to$adapted & \textbf{67$\to$52} & 284$\to$297 & 49$\to$56 & 400$\to$405 \\
\bottomrule
\end{tabular}
\end{table}

\section{Analysis and Diagnostic Controls}
\label{sec:controls}

\textbf{Judge substitution.} Most comparisons survive a change of judge. Rescoring every GPT-5.1-judged frontier turn with Sonnet 4.6 reproduces 90.4\% of decisions, while absolute rates rise by 2.9 to 4.8 points, especially for rubric-only IF items. \docmode stays above \ragmode for both readers, and only the close \ragmode ordering of GPT and Sonnet reverses. The switch widens GPT's lead over Sonnet by 1.3 points (persona-bootstrap 95\% CI $[-0.6,+3.3]$); we detect no preferential credit for GPT answers from the GPT judge, and the length interaction keeps its size (Appendix Table~\ref{tab:length-interaction}). Blinded human review of 24 stratified judge disagreements found each sampled gold label and rubric valid. This sample of disputed items gives neither a corpus-wide accuracy estimate nor a human upper bound.

\textbf{Second authoring family.} On an exploratory 16-persona panel written by Sonnet 4.6 and scored without any GPT model (260 rated turns), GPT-5.1 still outscores Sonnet under \docmode (216 against 198; 9 personas favor GPT and none Sonnet, sign-flip $p=0.004$). With 550-character excerpts, the gap changes from 1k to 10k by $-50.5$ (GPT, $[-67.8,-21.2]$) and $-35.1$ points (Sonnet, $[-66.9,+20.2]$); with whole sections, by $-6.2$ and $+7.2$, both intervals including zero (Appendix~\ref{app:sonnet-panel}).

\textbf{Item check.} Two validators (GPT-5.1, Sonnet 4.6) that see each dossier, SF question, and key, but no reader answer, flag 19 of 93 rated SF items (66 of 360 SF turns). A flag means either validator finds less than full support, including partial support and judge disagreement. Flagged turns pass 80.3\% under \docmode against 97.3\% and 94.6\% for the rest, and the SF length interaction holds without them (Appendix~\ref{app:item-check}).

\textbf{Planner substitution and regrading.} Aggregate frontier rates are stable across planners: a fixed heuristic planner changes frontier \ragmode accuracy by $-2.5$ (GPT) and $-4.6$ points (Sonnet), and both intervals include zero (Appendix~\ref{app:supp-replay}); the heuristic control also used 450-character excerpts and a 0.18 score floor, so this contrast changes excerpt length together with the planner. Under SF$_\text{key}$, the length interaction does not differ between the two planners ($+0.7$ and $+1.1$ points; both intervals include zero). Judge decisions are not fully deterministic: regrading 154 byte-identical responses changed 6 decisions, and the Section~\ref{sec:item-family} contrasts cache judge calls.

\textbf{Answer length.} Frontier answers were limited to 120 tokens, and the runner retries IF and NIP answers that appear truncated with larger limits. Under \ragmode, 0 (GPT) and 1 (Sonnet) of 360 SF answers are flagged as potentially truncated, against 4 and 14 under \docmode. We regenerated the 101 flagged frontier answers from identical message histories with a 1,200-token limit. Against the same answers re-judged in the same run, the regenerated answers add at most 6 passes to a row of Table~\ref{tab:replay}, almost all on IF; the access gaps move from $-8.5$ to $-9.0$ (GPT) and stay at $-4.0$ points (Sonnet), and the length interaction changes by $-2.9$ points for GPT ($[-5.8,-1.0]$) and not for Sonnet (Appendix Table~\ref{tab:cap-rerun}).

\textbf{Failure attribution.} The strict key rule requires the selected labeled sections to satisfy the key. Among 128 frontier SF failures it assigns 19 to retrieval, 47 to packaging, 39 to answer formation, and 23 to labeled sources lacking key support (Appendix Table~\ref{tab:attribution}). Applying GPT-5.1 and Sonnet 4.6 to the same evidence unions without keys yields 11 jointly assigned retrieval failures and 17 label disagreements (Appendix Table~\ref{tab:aligned-attribution}). Both judges give the packaging label to 17 of 21 GPT and 21 of 26 Sonnet key-based packaging cases. In the matched excerpt control, whole sections repair 29 of 32 and 25 of 30 packaging failures, compared with 2 of 8 and 1 of 8 answer-formation failures. For 19 of 21 and 24 of 26 Table~\ref{tab:replay} packaging failures, the key appears in neither the profile nor earlier turns; the item check flags 6 of 21 and 8 of 26 packaging cases (Appendix Table~\ref{tab:context-check}). The counts describe evidence availability under each support measure. Attribution remains an operational diagnostic with model disagreement; this audit has no independent human reference.

\section{Limitations and Scope}

\textbf{Judge reliability.} Model judges score document quality and part of replay; exact-tier agreement remains below within-one-tier agreement, and absolute replay rates shift under substitution (Section~\ref{sec:controls}). The SF entailment prompt changed between runs, so cross-run SF comparisons rely on the answer key.

\textbf{Scope of claims.} The cohort spans seven English-language generators and four length buckets. Main replay uses one source generator, one retriever family, and one decoding setting per target; replay dossiers and items are written by GPT-5.1, and the Sonnet-written panel is small and was extended from 8 to 16 personas after a pilot; frontier readers ran without extended reasoning; each reader row is a pipeline condition tied to its source records and does not rank generators. The format and adaptation contrasts use one open-source reader, chosen for consistent performance (Section~\ref{sec:extrinsic}).

\textbf{Framework assumptions.} The 11 seed axes and phase rules are incomplete. Embedding thresholds lack universal collapse criteria. Model-authored replay items and rubrics follow a fixed schema, and item authoring does not check each key against the dossier (Section~\ref{sec:controls}); freezing blocks target-output adaptation but retains authoring-model dependence. Deterministic rules can under-credit faithful paraphrases; rubric-only items remain judge-sensitive. Trait and diversity audits establish neither psychological validity nor population representativeness of the personas.

\textbf{Causal interpretation.} The payload study uses matched contextual-stress prompts and detects exact cross-scope anchors. Paraphrased borrowing can escape that test and the deterministic packet audit, and some labeled sections fail their frozen key. Each persona is written at one length, leaving length confounded with persona and question composition. A matched study of the same states and questions is needed. Eight-persona panels are too small for stable length or episode-type estimates. The adaptation contrast covers one adapter recipe on 32 personas.

\section{Conclusion}

\datasetname links intended state, generated record, delivered evidence, and reader answer. These links support offline tests of memory construction and use against known facts. Tag-scoped inputs reduce detected out-of-scope anchor copying in a contextual-stress control, and staged-authored questions transfer less reliably to one-pass records of the same states. Evidence controls show which packaging-labeled failures improve when whole sections are delivered. Attribution uses separate key-based and key-free checks, with disagreements retained. The between-bucket length result depends on excerpt construction; an independent length effect remains unresolved. For Gemma, adaptation lowers IF accuracy while aggregate changes remain inconclusive. A larger matched length study, more source generators, and independent human support annotation would test the remaining limits.

\subsection*{AI use statement}
In this work, we used generative AI tools for the following tasks with required disclosure. Large language models are part of the method: they generate the synthetic data sets (the dossiers and cards written by the generators of Section~\ref{sec:construction} and the frozen replay items authored by GPT-5.1), answer the replay questions, annotate the adaptation data, and support the analysis as judges, with identifiers and settings in Appendix Table~\ref{tab:config}. Separately, the authors used AI assistants under their direction and review to modify manuscript text (wording, restructuring, formatting, and consistency checks) and to implement and revise analysis and release code to the authors' specifications. During drafting, they refined the language of author-written passages for clarity, concision, and consistent terminology. These assistants also reformatted run outputs into the redacted release traces and gave feedback on experimental design and the interpretation of results. We have not used generative AI tools to propose the research questions or hypotheses: the authors defined them and the study design, launched every experiment, and decided which results and claims to report. The remaining tasks with required disclosure, mathematical claims and proofs and translation, are not applicable. We have reviewed all AI-assisted work: no AI-assisted text or code entered the paper without author review, every reported number traces to a run artifact or analysis output, and the release's validation script recomputes the replay numbers from the redacted traces. We take responsibility for the final content of this work, including text, claims, and artifacts produced with the aid of generative AI.

\subsection*{Ethics statement}
Synthetic personas can reproduce stereotypes, unrealistic correlations, and false impressions of completeness \citep{bender2018datastatements,gebru2021datasheets}. \datasetname provides no evidence about real populations and should not support high-stakes personalization, screening, or policy. Released artifacts need checks for memorized personal data and documented handling.

\subsection*{Reproducibility statement}
The anonymous repository contains the released persona cohort, construction-audit and calibration records, redacted turn-level traces for every reported replay condition, analysis outputs, and the code for construction, replay, encoding, and adaptation. A validation script recomputes the reported replay numbers from the traces. Model identifiers, decoding and retrieval settings, and seeds are recorded in Appendix Table~\ref{tab:config}, which also discloses historical configuration gaps. Replay prompts, answer keys, and completions are withheld to keep the frozen items usable as a held-out test. The repository is available at \url{https://anonymous.4open.science/r/ICLR_CHRONODOSSIER-4BCF}.

\bibliography{custom,iclr_additions,representation_additions}

\begin{thebibliography}{41}
\providecommand{\natexlab}[1]{#1}
\providecommand{\url}[1]{\texttt{#1}}
\expandafter\ifx\csname urlstyle\endcsname\relax
  \providecommand{\doi}[1]{doi: #1}\else
  \providecommand{\doi}{doi: \begingroup \urlstyle{rm}\Url}\fi

\bibitem[Ahn et~al.(2024)Ahn, Lee, Lim, Kim, Yun, Lee, and
  Kim]{ahn2024timechara}
Jaewoo Ahn, Taehyun Lee, Junyoung Lim, Jin-Hwa Kim, Sangdoo Yun, Hwaran Lee,
  and Gunhee Kim.
\newblock {TimeChara}: Evaluating point-in-time character hallucination of
  role-playing large language models.
\newblock In \emph{Findings of the Association for Computational Linguistics:
  ACL 2024}, pp.\  3291--3325, Bangkok, Thailand, 2024. Association for
  Computational Linguistics.
\newblock \doi{10.18653/v1/2024.findings-acl.197}.
\newblock URL \url{https://aclanthology.org/2024.findings-acl.197/}.

\bibitem[Bender \& Friedman(2018)Bender and Friedman]{bender2018datastatements}
Emily~M. Bender and Batya Friedman.
\newblock Data statements for natural language processing: Toward mitigating
  system bias and enabling better science.
\newblock \emph{Transactions of the Association for Computational Linguistics},
  6:\penalty0 587--604, 2018.
\newblock \doi{10.1162/tacl_a_00041}.
\newblock URL \url{https://aclanthology.org/Q18-1041/}.

\bibitem[Cao et~al.(2026)Cao, Ouyang, Cheng, Zhou, Liu, Li, Liu, and
  Wang]{cao2026re3}
Jiawei Cao, Jie Ouyang, Mingyue Cheng, Zhaomeng Zhou, Chunli Liu, Yupeng Li,
  Zirui Liu, and Shijin Wang.
\newblock Re$^3$: Relevance {\&} recency retrieval for mitigating temporal
  hallucination.
\newblock In \emph{Proceedings of the 64th Annual Meeting of the Association
  for Computational Linguistics (Volume 1: Long Papers)}, pp.\  25735--25760,
  San Diego, California, United States, 2026. Association for Computational
  Linguistics.
\newblock \doi{10.18653/v1/2026.acl-long.1180}.
\newblock URL \url{https://aclanthology.org/2026.acl-long.1180/}.

\bibitem[Chen et~al.(2024)Chen, Wang, Chen, Yu, Ma, Zhao, Zhang, and
  Yu]{chen2024densex}
Tong Chen, Hongwei Wang, Sihao Chen, Wenhao Yu, Kaixin Ma, Xinran Zhao,
  Hongming Zhang, and Dong Yu.
\newblock Dense {X} retrieval: What retrieval granularity should we use?
\newblock In \emph{Proceedings of the 2024 Conference on Empirical Methods in
  Natural Language Processing}, pp.\  15159--15177, Miami, Florida, USA, 2024.
  Association for Computational Linguistics.
\newblock URL \url{https://aclanthology.org/2024.emnlp-main.845/}.

\bibitem[Chen et~al.(2025)Chen, Min, Zhao, Li, Jia, Liao, Li, Wang, Hu, and
  Yin]{chen2025chronoqa}
Ziyang Chen, Erxue Min, Xiang Zhao, Yunxin Li, Xin Jia, Jinzhi Liao, Jichao Li,
  Shuaiqiang Wang, Baotian Hu, and Dawei Yin.
\newblock A question answering dataset for temporal-sensitive
  retrieval-augmented generation.
\newblock \emph{Scientific Data}, 12\penalty0 (1):\penalty0 1855, 2025.
\newblock \doi{10.1038/s41597-025-06098-y}.
\newblock URL \url{https://doi.org/10.1038/s41597-025-06098-y}.

\bibitem[Cheng et~al.(2026)Cheng, Ru, Qiu, Li, Cao, Song, and
  Cai]{cheng2026amemgym}
Jiayang Cheng, Dongyu Ru, Lin Qiu, Yiyang Li, Xuezhi Cao, Yangqiu Song, and
  Xunliang Cai.
\newblock {AMemGym}: Interactive memory benchmarking for assistants in
  long-horizon conversations.
\newblock In \emph{International Conference on Learning Representations}, 2026.
\newblock URL
  \url{https://proceedings.iclr.cc/paper_files/paper/2026/hash/0856bc553d3e3b9827e5140d0ad3bf8d-Abstract-Conference.html}.

\bibitem[Dettmers et~al.(2023)Dettmers, Pagnoni, Holtzman, and
  Zettlemoyer]{dettmers2023qlora}
Tim Dettmers, Artidoro Pagnoni, Ari Holtzman, and Luke Zettlemoyer.
\newblock {QLoRA}: Efficient finetuning of quantized {LLMs}.
\newblock In \emph{Advances in Neural Information Processing Systems},
  volume~36, 2023.
\newblock URL
  \url{https://proceedings.neurips.cc/paper_files/paper/2023/hash/1feb87871436031bdc0f2beaa62a049b-Abstract-Conference.html}.

\bibitem[Fan et~al.(2026)Fan, Chen, Hu, Peng, Huang, Xu, Chen, Wu, Zhao, and
  Liu]{fan2026optimsyn}
Zhiting Fan, Ruizhe Chen, Tianxiang Hu, Ru~Peng, Zenan Huang, Haokai Xu, Yixin
  Chen, Jian Wu, Junbo Zhao, and Zuozhu Liu.
\newblock {OptimSyn}: Influence-guided rubrics optimization for synthetic data
  generation.
\newblock In \emph{International Conference on Learning Representations}, 2026.
\newblock URL
  \url{https://proceedings.iclr.cc/paper_files/paper/2026/hash/b0051386fa004515c1fc3cdda8080b04-Abstract-Conference.html}.

\bibitem[Ge et~al.(2024)Ge, Chan, Wang, Yu, Mi, and Yu]{ge2024personahub}
Tao Ge, Xin Chan, Xiaoyang Wang, Dian Yu, Haitao Mi, and Dong Yu.
\newblock Scaling synthetic data creation with 1,000,000,000 personas, 2024.
\newblock URL \url{https://arxiv.org/abs/2406.20094}.
\newblock arXiv:2406.20094.

\bibitem[Gebru et~al.(2021)Gebru, Morgenstern, Vecchione, Wortman~Vaughan,
  Wallach, Daum{\'e}~III, and Crawford]{gebru2021datasheets}
Timnit Gebru, Jamie Morgenstern, Briana Vecchione, Jennifer Wortman~Vaughan,
  Hanna Wallach, Hal Daum{\'e}~III, and Kate Crawford.
\newblock Datasheets for datasets.
\newblock \emph{Communications of the ACM}, 64\penalty0 (12):\penalty0 86--92,
  2021.
\newblock URL \url{https://cacm.acm.org/research/datasheets-for-datasets/}.

\bibitem[{Gemma Team}(2026)]{gemmateam2026gemma4}
{Gemma Team}.
\newblock {Gemma 4} technical report, 2026.
\newblock URL \url{https://arxiv.org/abs/2607.02770}.
\newblock Technical report, arXiv:2607.02770.

\bibitem[Grattafiori et~al.(2024)]{grattafiori2024llama3}
Aaron Grattafiori et~al.
\newblock The {Llama} 3 herd of models, 2024.
\newblock URL \url{https://arxiv.org/abs/2407.21783}.
\newblock Technical report, arXiv:2407.21783.

\bibitem[Hu et~al.(2022)Hu, Shen, Wallis, Allen-Zhu, Li, Wang, Wang, and
  Chen]{hu2022lora}
Edward~J. Hu, Yelong Shen, Phillip Wallis, Zeyuan Allen-Zhu, Yuanzhi Li, Shean
  Wang, Lu~Wang, and Weizhu Chen.
\newblock {LoRA}: Low-rank adaptation of large language models.
\newblock In \emph{International Conference on Learning Representations}, 2022.
\newblock URL \url{https://openreview.net/forum?id=nZeVKeeFYf9}.

\bibitem[Jiang et~al.(2025)Jiang, Hao, Cho, Li, Yuan, Chen, Ungar, Taylor, and
  Roth]{jiang2025personamem}
Bowen Jiang, Zhuoqun Hao, Young-Min Cho, Bryan Li, Yuan Yuan, Sihao Chen, Lyle
  Ungar, Camillo~J. Taylor, and Dan Roth.
\newblock Know me, respond to me: Benchmarking {LLMs} for dynamic user
  profiling and personalized responses at scale.
\newblock In \emph{Conference on Language Modeling}, 2025.
\newblock URL \url{https://arxiv.org/abs/2504.14225}.

\bibitem[Lewis et~al.(2020)Lewis, Perez, Piktus, Petroni, Karpukhin, Goyal,
  Kuttler, Lewis, Yih, Rocktaschel, Riedel, and Kiela]{lewis2020rag}
Patrick Lewis, Ethan Perez, Aleksandra Piktus, Fabio Petroni, Vladimir
  Karpukhin, Naman Goyal, Heinrich Kuttler, Mike Lewis, Wen-tau Yih, Tim
  Rocktaschel, Sebastian Riedel, and Douwe Kiela.
\newblock Retrieval-augmented generation for knowledge-intensive {NLP} tasks.
\newblock In \emph{Advances in Neural Information Processing Systems},
  volume~33, pp.\  9459--9474, 2020.
\newblock URL
  \url{https://papers.nips.cc/paper/2020/hash/6b493230205f780e1bc26945df7481e5-Abstract.html}.

\bibitem[Liu et~al.(2024)Liu, Lin, Hewitt, Paranjape, Bevilacqua, Petroni, and
  Liang]{liu2024lost}
Nelson~F. Liu, Kevin Lin, John Hewitt, Ashwin Paranjape, Michele Bevilacqua,
  Fabio Petroni, and Percy Liang.
\newblock Lost in the middle: How language models use long contexts.
\newblock \emph{Transactions of the Association for Computational Linguistics},
  12:\penalty0 157--173, 2024.
\newblock \doi{10.1162/tacl_a_00638}.
\newblock URL \url{https://aclanthology.org/2024.tacl-1.9/}.

\bibitem[Liu et~al.(2023)Liu, Iter, Xu, Wang, Xu, and Zhu]{liu2023geval}
Yang Liu, Dan Iter, Yichong Xu, Shuohang Wang, Ruochen Xu, and Chenguang Zhu.
\newblock {G}-eval: {NLG} evaluation using {GPT-4} with better human alignment.
\newblock In \emph{Proceedings of the 2023 Conference on Empirical Methods in
  Natural Language Processing}, pp.\  2511--2522, 2023.
\newblock \doi{10.18653/v1/2023.emnlp-main.153}.
\newblock URL \url{https://aclanthology.org/2023.emnlp-main.153/}.

\bibitem[Luo \& Laban(2026)Luo and Laban]{luo2026spasm}
Han Luo and Guy Laban.
\newblock {SPASM}: Stable persona-driven agent simulation for multi-turn
  dialogue generation.
\newblock In \emph{Findings of the Association for Computational Linguistics:
  ACL 2026}, pp.\  8455--8475, San Diego, California, United States, 2026.
  Association for Computational Linguistics.
\newblock \doi{10.18653/v1/2026.findings-acl.412}.
\newblock URL \url{https://aclanthology.org/2026.findings-acl.412/}.

\bibitem[Madotto et~al.(2019)Madotto, Lin, Wu, and
  Fung]{madotto2019personalizing}
Andrea Madotto, Zhaojiang Lin, Chien-Sheng Wu, and Pascale Fung.
\newblock Personalizing dialogue agents via meta-learning.
\newblock In \emph{Proceedings of the 57th Annual Meeting of the Association
  for Computational Linguistics}, pp.\  5454--5459, Florence, Italy, 2019.
  Association for Computational Linguistics.
\newblock \doi{10.18653/v1/P19-1542}.
\newblock URL \url{https://aclanthology.org/P19-1542/}.

\bibitem[Maharana et~al.(2024)Maharana, Lee, Tulyakov, Bansal, Barbieri, and
  Fang]{maharana2024locomo}
Adyasha Maharana, Dong-Ho Lee, Sergey Tulyakov, Mohit Bansal, Francesco
  Barbieri, and Yuwei Fang.
\newblock Evaluating very long-term conversational memory of {LLM} agents.
\newblock In \emph{Proceedings of the 62nd Annual Meeting of the Association
  for Computational Linguistics (Volume 1: Long Papers)}, pp.\  13851--13870,
  Bangkok, Thailand, 2024. Association for Computational Linguistics.
\newblock \doi{10.18653/v1/2024.acl-long.747}.

\bibitem[Mitchell et~al.(2019)Mitchell, Wu, Zaldivar, Barnes, Vasserman,
  Hutchinson, Spitzer, Raji, and Gebru]{mitchell2019modelcards}
Margaret Mitchell, Simone Wu, Andrew Zaldivar, Parker Barnes, Lucy Vasserman,
  Ben Hutchinson, Elena Spitzer, Inioluwa~Deborah Raji, and Timnit Gebru.
\newblock Model cards for model reporting.
\newblock In \emph{Proceedings of the Conference on Fairness, Accountability,
  and Transparency}, pp.\  220--229, 2019.
\newblock \doi{10.1145/3287560.3287596}.
\newblock URL \url{https://doi.org/10.1145/3287560.3287596}.

\bibitem[Packer et~al.(2023)Packer, Wooders, Lin, Fang, Patil, Stoica, and
  Gonzalez]{packer2023memgpt}
Charles Packer, Sarah Wooders, Kevin Lin, Vivian Fang, Shishir~G. Patil, Ion
  Stoica, and Joseph~E. Gonzalez.
\newblock {MemGPT}: Towards {LLMs} as operating systems, 2023.
\newblock URL \url{https://arxiv.org/abs/2310.08560}.
\newblock arXiv:2310.08560.

\bibitem[Pan et~al.(2025)Pan, Wu, Jiang, Luo, Cheng, Li, Yang, Lin, Zhao, Qiu,
  and Gao]{pan2025secom}
Zhuoshi Pan, Qianhui Wu, Huiqiang Jiang, Xufang Luo, Hao Cheng, Dongsheng Li,
  Yuqing Yang, Chin-Yew Lin, H.~Vicky Zhao, Lili Qiu, and Jianfeng Gao.
\newblock {SeCom}: On memory construction and retrieval for personalized
  conversational agents.
\newblock In \emph{International Conference on Learning Representations}, 2025.
\newblock URL
  \url{https://proceedings.iclr.cc/paper_files/paper/2025/hash/e56f394bbd4f0ec81393d767caa5a31b-Abstract-Conference.html}.

\bibitem[Park et~al.(2023)Park, O'Brien, Cai, Morris, Liang, and
  Bernstein]{park2023generative}
Joon~Sung Park, Joseph~C. O'Brien, Carrie~J. Cai, Meredith~Ringel Morris, Percy
  Liang, and Michael~S. Bernstein.
\newblock Generative agents: Interactive simulacra of human behavior.
\newblock In \emph{Proceedings of the 36th Annual ACM Symposium on User
  Interface Software and Technology (UIST '23)}. ACM, 2023.
\newblock URL \url{https://arxiv.org/abs/2304.03442}.

\bibitem[{Qwen Team}(2025)]{qwen2025report}
{Qwen Team}.
\newblock {Qwen3} technical report, 2025.
\newblock URL \url{https://arxiv.org/abs/2505.09388}.
\newblock Technical report, arXiv:2505.09388.

\bibitem[Reimers \& Gurevych(2019)Reimers and
  Gurevych]{reimers2019sentencebert}
Nils Reimers and Iryna Gurevych.
\newblock Sentence-{BERT}: Sentence embeddings using {Siamese} {BERT}-networks.
\newblock In \emph{Proceedings of the 2019 Conference on Empirical Methods in
  Natural Language Processing and the 9th International Joint Conference on
  Natural Language Processing}, pp.\  3982--3992, 2019.
\newblock \doi{10.18653/v1/D19-1410}.
\newblock URL \url{https://aclanthology.org/D19-1410/}.

\bibitem[Samuel et~al.(2025)Samuel, Zou, Zhou, Chaudhari, Kalyan, Rajpurohit,
  Deshpande, Narasimhan, and Murahari]{samuel2024personagym}
Vinay Samuel, Henry~Peng Zou, Yue Zhou, Shreyas Chaudhari, Ashwin Kalyan,
  Tanmay Rajpurohit, Ameet Deshpande, Karthik~R. Narasimhan, and Vishvak
  Murahari.
\newblock {PersonaGym}: Evaluating persona agents and {LLM}s.
\newblock In \emph{Findings of the Association for Computational Linguistics:
  EMNLP 2025}, pp.\  6999--7022, Suzhou, China, 2025. Association for
  Computational Linguistics.
\newblock \doi{10.18653/v1/2025.findings-emnlp.368}.
\newblock URL \url{https://aclanthology.org/2025.findings-emnlp.368/}.

\bibitem[Shi et~al.(2026)Shi, Lian, Wu, Qin, Zhou, Xie, Chao, and
  Liao]{shi2026personaarena}
Wenlong Shi, Jianxun Lian, Mingqi Wu, Haiming Qin, Mingyang Zhou, Xing Xie,
  Naipeng Chao, and Hao Liao.
\newblock {PersonaArena}: Dynamic simulation for evaluating and enhancing
  persona-level role-playing in large language models.
\newblock In \emph{Findings of the Association for Computational Linguistics:
  ACL 2026}, pp.\  9685--9719, San Diego, California, United States, 2026.
  Association for Computational Linguistics.
\newblock \doi{10.18653/v1/2026.findings-acl.471}.
\newblock URL \url{https://aclanthology.org/2026.findings-acl.471/}.

\bibitem[Tong \& Zou(2026)Tong and Zou]{tong2026personaforge}
Jizhou Tong and Sirui Zou.
\newblock {PersonaForge}: Psychology-grounded dual-process architecture for
  personality-consistent role-playing agents.
\newblock In \emph{Findings of the Association for Computational Linguistics:
  ACL 2026}, pp.\  7845--7874, San Diego, California, United States, 2026.
  Association for Computational Linguistics.
\newblock \doi{10.18653/v1/2026.findings-acl.386}.
\newblock URL \url{https://aclanthology.org/2026.findings-acl.386/}.

\bibitem[Wang et~al.(2024{\natexlab{a}})Wang, Peng, Que, Liu, Zhou, Wu, Guo,
  Gan, Ni, Yang, Zhang, Zhang, Ouyang, Xu, Huang, Fu, and
  Peng]{wang2024rolellm}
Noah Wang, Z.Y. Peng, Haoran Que, Jiaheng Liu, Wangchunshu Zhou, Yuhan Wu,
  Hongcheng Guo, Ruitong Gan, Zehao Ni, Jian Yang, Man Zhang, Zhaoxiang Zhang,
  Wanli Ouyang, Ke~Xu, Wenhao Huang, Jie Fu, and Junran Peng.
\newblock {RoleLLM}: Benchmarking, eliciting, and enhancing role-playing
  abilities of large language models.
\newblock In \emph{Findings of the Association for Computational Linguistics:
  ACL 2024}, pp.\  14743--14777, Bangkok, Thailand, 2024{\natexlab{a}}.
  Association for Computational Linguistics.
\newblock \doi{10.18653/v1/2024.findings-acl.878}.
\newblock URL \url{https://aclanthology.org/2024.findings-acl.878/}.

\bibitem[Wang et~al.(2025)Wang, Zhang, Ge, Yu, Yu, and
  Yu]{wang2025opencharacter}
Xiaoyang Wang, Hongming Zhang, Tao Ge, Wenhao Yu, Dian Yu, and Dong Yu.
\newblock {OpenCharacter}: Training customizable role-playing llms with
  large-scale synthetic personas, 2025.
\newblock URL \url{https://arxiv.org/abs/2501.15427}.
\newblock arXiv:2501.15427.

\bibitem[Wang et~al.(2024{\natexlab{b}})Wang, Xiao, Huang, Yuan, Xu, Guo, Tu,
  Fei, Leng, Wang, Chen, Li, and Xiao]{wang2024incharacter}
Xintao Wang, Yunze Xiao, Jen-tse Huang, Siyu Yuan, Rui Xu, Haoran Guo, Quan Tu,
  Yaying Fei, Ziang Leng, Wei Wang, Jiangjie Chen, Cheng Li, and Yanghua Xiao.
\newblock {InCharacter}: Evaluating personality fidelity in role-playing agents
  through psychological interviews.
\newblock In \emph{Proceedings of the 62nd Annual Meeting of the Association
  for Computational Linguistics (Volume 1: Long Papers)}, pp.\  1840--1873,
  Bangkok, Thailand, 2024{\natexlab{b}}. Association for Computational
  Linguistics.
\newblock \doi{10.18653/v1/2024.acl-long.102}.
\newblock URL \url{https://aclanthology.org/2024.acl-long.102/}.

\bibitem[Welleck et~al.(2019)Welleck, Weston, Szlam, and
  Cho]{welleck2019dialoguenli}
Sean Welleck, Jason Weston, Arthur Szlam, and Kyunghyun Cho.
\newblock Dialogue natural language inference.
\newblock In \emph{Proceedings of the 57th Annual Meeting of the Association
  for Computational Linguistics}, pp.\  3731--3741, 2019.
\newblock \doi{10.18653/v1/P19-1363}.
\newblock URL \url{https://aclanthology.org/P19-1363/}.

\bibitem[Wu et~al.(2025)Wu, Wang, Yu, Zhang, Chang, and Yu]{wu2025longmemeval}
Di~Wu, Hongwei Wang, Wenhao Yu, Yuwei Zhang, Kai-Wei Chang, and Dong Yu.
\newblock {LongMemEval}: Benchmarking chat assistants on long-term interactive
  memory.
\newblock In \emph{The Thirteenth International Conference on Learning
  Representations (ICLR)}, Singapore, 2025.
\newblock URL \url{https://openreview.net/forum?id=pZiyCaVuti}.

\bibitem[Xu et~al.(2022)Xu, Szlam, and Weston]{xu2022goldfish}
Jing Xu, Arthur Szlam, and Jason Weston.
\newblock Beyond goldfish memory: Long-term open-domain conversation.
\newblock In \emph{Proceedings of the 60th Annual Meeting of the Association
  for Computational Linguistics (Volume 1: Long Papers)}, pp.\  5180--5197,
  Dublin, Ireland, 2022. Association for Computational Linguistics.
\newblock \doi{10.18653/v1/2022.acl-long.356}.
\newblock URL \url{https://aclanthology.org/2022.acl-long.356/}.

\bibitem[Yen et~al.(2025)Yen, Gao, Hou, Ding, Fleischer, Izsak, Wasserblat, and
  Chen]{yen2025helmet}
Howard Yen, Tianyu Gao, Minmin Hou, Ke~Ding, Daniel Fleischer, Peter Izsak,
  Moshe Wasserblat, and Danqi Chen.
\newblock {HELMET}: How to evaluate long-context models effectively and
  thoroughly.
\newblock In \emph{International Conference on Learning Representations}, 2025.
\newblock URL
  \url{https://proceedings.iclr.cc/paper_files/paper/2025/hash/f5332c8273d02729730a9c24dec2135e-Abstract-Conference.html}.

\bibitem[Zhang et~al.(2018)Zhang, Dinan, Urbanek, Szlam, Kiela, and
  Weston]{zhang2018persona}
Saizheng Zhang, Emily Dinan, Jack Urbanek, Arthur Szlam, Douwe Kiela, and Jason
  Weston.
\newblock Personalizing dialogue agents: I have a dog, do you have pets too?
\newblock In \emph{Proceedings of the 56th Annual Meeting of the Association
  for Computational Linguistics (Volume 1: Long Papers)}, pp.\  2204--2213,
  Melbourne, Australia, 2018. Association for Computational Linguistics.
\newblock \doi{10.18653/v1/P18-1205}.
\newblock URL \url{https://aclanthology.org/P18-1205/}.

\bibitem[Zhao et~al.(2025)Zhao, Hong, Liu, Hazarika, and Lin]{zhao2025prefeval}
Siyan Zhao, Mingyi Hong, Yang Liu, Devamanyu Hazarika, and Kaixiang Lin.
\newblock Do {LLMs} recognize your preferences? evaluating personalized
  preference following in {LLMs}.
\newblock In \emph{International Conference on Learning Representations}, 2025.
\newblock URL \url{https://openreview.net/forum?id=QWunLKbBGF}.

\bibitem[Zheng et~al.(2023)Zheng, Chiang, Sheng, Zhuang, Wu, Zhuang, Lin, Li,
  Li, Xing, Zhang, Gonzalez, and Stoica]{zheng2023judging}
Lianmin Zheng, Wei-Lin Chiang, Ying Sheng, Siyuan Zhuang, Zhanghao Wu, Yonghao
  Zhuang, Zi~Lin, Zhuohan Li, Dacheng Li, Eric~P. Xing, Hao Zhang, Joseph~E.
  Gonzalez, and Ion Stoica.
\newblock Judging {LLM}-as-a-judge with {MT-Bench} and chatbot arena.
\newblock In \emph{Advances in Neural Information Processing Systems 36:
  Datasets and Benchmarks Track}, 2023.
\newblock URL \url{https://openreview.net/forum?id=uccHPGDlao}.

\bibitem[Zhou et~al.(2025)Zhou, Huang, Wen, Bi, Chen, Ke, Chen, Xiao, Peng,
  Tang, Zhang, Zhang, Lv, Hu, Wang, and Huang]{zhou2025characterbench}
Jinfeng Zhou, Yongkang Huang, Bosi Wen, Guanqun Bi, Yuxuan Chen, Pei Ke, Zhuang
  Chen, Xiyao Xiao, Libiao Peng, Kuntian Tang, Rongsheng Zhang, Le~Zhang,
  Tangjie Lv, Zhipeng Hu, Hongning Wang, and Minlie Huang.
\newblock {CharacterBench}: Benchmarking character customization of large
  language models.
\newblock In \emph{Proceedings of the AAAI Conference on Artificial
  Intelligence}, volume~39, pp.\  26101--26110, 2025.
\newblock \doi{10.1609/aaai.v39i24.34806}.
\newblock URL \url{https://ojs.aaai.org/index.php/AAAI/article/view/34806}.

\bibitem[Zhou et~al.(2023)Zhou, Li, and Li]{zhou2023predictpersona}
Wangchunshu Zhou, Qifei Li, and Chenle Li.
\newblock Learning to predict persona information for dialogue personalization
  without explicit persona description.
\newblock In \emph{Findings of the Association for Computational Linguistics:
  ACL 2023}, pp.\  2979--2991, Toronto, Canada, 2023. Association for
  Computational Linguistics.
\newblock \doi{10.18653/v1/2023.findings-acl.186}.
\newblock URL \url{https://aclanthology.org/2023.findings-acl.186/}.

\end{thebibliography}
\bibliographystyle{iclr2027_conference}

\appendix
\FloatBarrier
\renewcommand{\topfraction}{0.9}
\renewcommand{\bottomfraction}{0.8}
\renewcommand{\textfraction}{0.07}
\renewcommand{\floatpagefraction}{0.85}
\setcounter{topnumber}{4}
\setcounter{bottomnumber}{3}
\setcounter{totalnumber}{6}
\section{Positioning Against Persona Resources}

\begin{table}[htbp]
\caption{Representative persona resources by representation scale and primary use. Units follow the source resources; medians are computed from the released profiles. The table positions research scope and does not rank native-task performance.}
\label{tab:positioning}
\centering
\small
\setlength{\tabcolsep}{3pt}
\renewcommand{\arraystretch}{1.05}
\begin{tabularx}{\textwidth}{@{}>{\raggedright\arraybackslash}p{0.24\textwidth}>{\raggedright\arraybackslash}p{0.39\textwidth}Y@{}}
\toprule
Resource & Representation & Primary use \\
\midrule
PERSONA-CHAT & Static profile; five sentences & Dialogue \\
Persona Hub & Web-derived descriptor; one string & Data synthesis \\
RoleLLM / RoleBench & Character profile; median 97 words & Role-play training and evaluation \\
PersonaGym & Static profile; median 10 words & Dynamic agent evaluation \\
CharacterBench & Character profile; median 106 words & Character customization \\
TimeChara & Narrative point plus QA item & Point-in-time fidelity \\
\rowcolor{black!5}\textbf{\datasetname} & \textbf{Year-bounded dossier; 1k--10k words} & \textbf{Construction audit and replay} \\
\bottomrule
\end{tabularx}
\end{table}

\FloatBarrier
\section{Artifact Schemas and Join Keys}
\label{app:schemas}

The released replay cohort contains five artifact-record families: dossier documents, compact cards, section records, phase capsules, and repair logs. The cohort manifest names each family and maps the four frontier replay rows to their stored run directories. The stable \texttt{run\_id} and \texttt{persona\_id} fields join persona-level records; \texttt{section\_id} joins retrieved passages to section records; phase and time-scope fields link sections with phase-local evidence. Frozen replay metadata preserves item identifiers, item family, and time metadata while withholding prompts, gold answers, rubrics, and model completions from the blinded bundle. Source artifacts: \path{data/replay_cohort/README.md}, \path{data/replay_cohort/manifest.json}, and \path{data/cited_artifacts/artifact_manifest.md}. Table~\ref{tab:artifact-schema} lists each released family with its row unit and join keys.

\begin{table}[htbp]
\caption{Released artifact schema and join keys. \texttt{persona\_id} links persona-level families; \texttt{run\_id} disambiguates generator runs; \texttt{section\_id}, \texttt{phase\_id}, \texttt{eval\_id}, and \texttt{episode\_id} identify progressively finer records.}
\label{tab:artifact-schema}
\centering
\small
\setlength{\tabcolsep}{3pt}
\renewcommand{\arraystretch}{1.1}
\begin{tabularx}{\textwidth}{@{}>{\raggedright\arraybackslash}p{0.31\textwidth}>{\raggedright\arraybackslash}p{0.11\textwidth}>{\raggedright\arraybackslash}p{0.23\textwidth}Y@{}}
\toprule
Artifact family & Row unit & Stable key / join & Contents and release scope \\
\midrule
\texttt{persona\_docs.jsonl} & accepted dossier & \texttt{run\_id}; \texttt{persona\_id} & Seed, timeline, outline, full dossier, length, and audit fields; all 311 released. \\
\texttt{persona\_cards.jsonl} & persona card & \texttt{run\_id}; \texttt{persona\_id} & Compact values, facts, tendencies, style, and boundaries; all 311 released. \\
\texttt{persona\_sections.jsonl} & dossier section & \texttt{run\_id}, \texttt{section\_id}; join \texttt{persona\_id} & Heading, tags, time scope, text, and generation and audit metadata; complete for the replay cohort. \\
\texttt{phase\_capsules.jsonl} & persona phase & \texttt{persona\_id}, \texttt{phase\_id}; \texttt{included\_}\newline\texttt{section\_ids} & Phase bounds and phase-local excerpts used by temporal audits; complete for the replay cohort. \\
\texttt{persona\_fix\_log.jsonl} & repair attempt & \texttt{run\_id}, \texttt{section\_id}, stage, attempt & Gate failures, actions, before and after metrics, and acceptance; complete for the replay cohort. \\
\texttt{frozen\_item\_}\newline\texttt{metadata.jsonl} & item definition & \texttt{eval\_id}; join \texttt{run\_id}, \texttt{persona\_id}, \texttt{phase\_id} & Family, subcategory, question shape, NLI relation, and time scope for 431 definitions; questions and gold text withheld during review. \\
\path{episode_results/*.jsonl} & model--mode episode & \texttt{episode\_id}, \texttt{session\_mode} & Aggregate turns, passes, history composition, retrieval counts, and usage for the four frontier rows; \path{traces/} covers every row. Turn text withheld during review. \\
\bottomrule
\end{tabularx}
\end{table}

\section{Construction Audit Details}
\label{app:construction-audits}

\begin{figure}[htbp]
\centering
\includegraphics[trim=115 214 135 150,clip,width=0.9\textwidth]{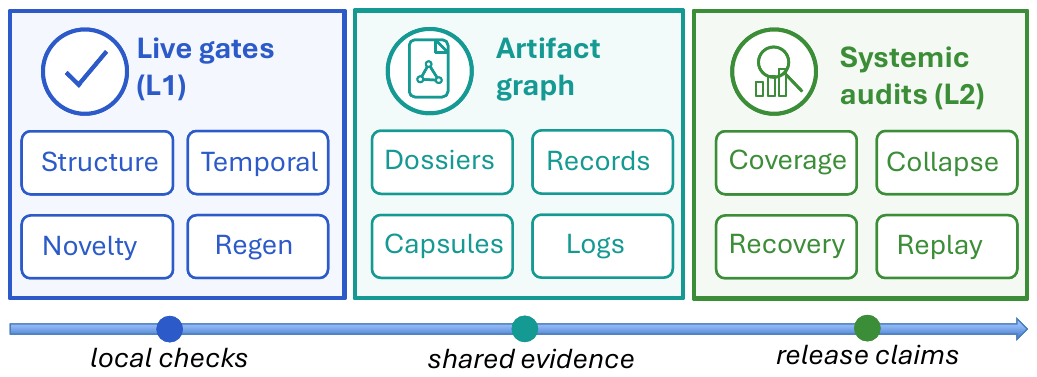}
\caption{Two-layer audit design, expanding the audit and replay stage of Figure~\ref{fig:framework}. Live gates (L1) enforce section-local admission and repair during generation; the exported artifact graph then feeds systemic audits (L2) of coverage, concentration, recoverability, and replay.}
\label{fig:audit-layers}
\end{figure}

Figure~\ref{fig:audit-layers} separates the live gates applied during generation from the systemic audits applied to exported records. Table~\ref{tab:rubric} lists the document-quality rubric shared by all judge providers.

\begin{table}[htbp]
\caption{Document-quality rubric criteria applied by every judge provider.}
\label{tab:rubric}
\centering
\small
\begin{tabular}{ll}
\toprule
Criterion & Purpose \\
\midrule
Internal consistency & Detect contradictions within and across sections. \\
Temporal consistency & Check adherence to the provided timeline and phase scopes. \\
Seed adherence & Verify faithfulness to the structured seed. \\
Specificity and realism & Reward concrete, plausible detail over generic prose. \\
Structure and readability & Evaluate organization and long-form readability. \\
Bias avoidance & Penalize stereotype-driven or overgeneralized content. \\
Retrieval readiness & Assess whether section structure supports retrieval use. \\
Length control & Check compliance with target length and section budgets. \\
\bottomrule
\end{tabular}
\end{table}

Table~\ref{tab:wiki-balance} compares matched synthetic and Wikipedia biographies. The synthetic documents are more concentrated in both embedding views, with comparable lexical 5-gram overlap. Table~\ref{tab:equal-size} removes cohort size as a source of nearest-neighbor opportunity; Qwen remains the most concentrated generator under equal-size subsampling.

\begin{table}[htbp]
\caption{Balanced synthetic-versus-Wikipedia concentration ($n=20$ per side). Wikipedia biographies are grounded, size-filtered, and bucket-matched by source length; the Wikipedia values are means over 21 subsets. Top1 and Pairwise are embedding similarities; 5-gram is lexical Jaccard overlap. Lower values indicate weaker concentration.}
\label{tab:wiki-balance}
\centering
\small
\begin{tabular}{lccc}
\toprule
Corpus & Top1$\downarrow$ & Pairwise$\downarrow$ & 5-gram$\downarrow$ \\
\midrule
Synthetic & 0.4039 & 0.2089 & 0.000166 \\
Wikipedia & 0.3727 & 0.1251 & 0.000210 \\
\bottomrule
\end{tabular}
\end{table}

\begin{table}[htbp]
\caption{Equal-size ($n=32$) concentration within each generator. Cells are medians and central 95\% intervals over 10,000 subsamples of 32 documents; Gemma has exactly 32 accepted documents and therefore one possible sample. Higher similarity indicates greater concentration.}
\label{tab:equal-size}
\centering
\small
\begin{tabular}{lrrr}
\toprule
Generator & $n$ & Top1 mean & Top1 p95 \\
\midrule
GPT-5.1 & 94 & .547 [.523, .572] & .643 [.585, .699] \\
GPT-5 mini & 37 & .575 [.564, .584] & .642 [.626, .642] \\
Sonnet 4.6 & 35 & .534 [.520, .542] & .612 [.584, .612] \\
Haiku 4.5 & 37 & .519 [.506, .529] & .609 [.584, .609] \\
Gemma 4 26B-A4B & 32 & .627 [.627, .627] & .689 [.689, .689] \\
Llama 3.1 70B & 37 & .565 [.550, .576] & .664 [.639, .664] \\
Qwen3 32B & 39 & \textbf{.772 [.760, .784]} & \textbf{.838 [.821, .838]} \\
\bottomrule
\end{tabular}
\end{table}

Table~\ref{tab:seed-axes} lists the values of the 11 seed axes and their realized distribution over the 311 accepted dossiers and the 32 replay personas. Region values are setting descriptors, such as ``large coastal city'', ``rural Appalachia'', and ``gulf coast city''; many are US-oriented, and the axis does not define a set of countries.

\begin{table}[htbp]
\caption{Seed axes. Values counts the distinct values of each axis, all of which occur among the 311 accepted dossiers; the range gives the fewest and most dossiers per value. The replay column gives the 32 replay personas.}
\label{tab:seed-axes}
\centering
\small
\setlength{\tabcolsep}{4pt}
\begin{tabular}{lrll}
\toprule
Axis & Values & Dossiers per value (311) & Replay panel (32) \\
\midrule
Age range & 5 & 61--64 & 5--8 per range \\
Gender & 3 & woman 152, man 143, other 16 & 15, 15, 2 \\
Ethnicity or background & 8 & 15--55 & all 8 values \\
Region (setting descriptor) & 31 & 7--16 & 24 of 31 \\
Urban form & 12 & 21--31 & all 12 \\
Climate context & 10 & 22--46 & all 10 \\
Education & 6 & 31--84 & all 6 \\
Wealth & 5 & 42--98 & all 5 \\
Job & 71 & 1--8 & 30 of 71 \\
Job specialization & 12 & 21--31 & all 12 \\
Work schedule & 12 & 21--31 & all 12 \\
\bottomrule
\end{tabular}
\end{table}

\subsection{Phase-rubric calibration}
\label{app:calibration}

GPT-5.1 agrees with the reference human nearly as often as the two humans agree with each other; the other two judges are harsher on disputed packets. On 25 shared phase packets, the primary human pair agrees exactly on 17/25 and within one tier on 25/25, and GPT-5.1 agrees with the reference human exactly on 16/25 and within one tier on 24/25. Gemini 3 Pro Preview and Claude Sonnet 4.5 agree exactly on 12/25 and 11/25 and within one tier on 19/25 and 20/25. Sonnet 4.5 was used because the calibration predates the main Sonnet 4.6 replay.

\section{Replay Configuration and Count Units}
\label{app:replay-config}

The frontier replay configuration uses target-model query planning, normalized all-MiniLM-L6-v2 embeddings, hybrid section retrieval with $k=3$, a minimum score of 0.08, a 550-character excerpt cap, and a two-turn query-history window. The 32 replay personas are a seeded sample (seed 20260423) of eight GPT-5.1 dossiers per length bucket; two personas whose item generation had returned malformed JSON were excluded before sampling. The other 62 GPT-5.1 dossiers are the training and development personas of the adaptation study. The count units nest as follows: 311 accepted dossiers, 94 from GPT-5.1, 32 replay personas, 431 frozen item definitions, 208 distinct rated items, and 520 rated turns per condition, within 96 episodes of 8,880 turns. Scheduled retention and time-jump recurrences turn the 208 items into 520 rated turns. These units remain separate in the main-text denominator labels. Source artifacts: \path{analysis/audit_and_replay/reproducibility_configuration.json}, \path{data/replay_cohort/README.md}, and \path{data/replay_cohort/manifest.json}.

The retained GPT and Sonnet frontier turn records report \texttt{retrieval\_embed\_model\_requested} and \texttt{retrieval\_embed\_model\_used} as \texttt{sentence-transformers/all-MiniLM-L6-v2} on rated turns. An earlier GPT run requested \texttt{Alibaba-NLP/gte-Qwen2-7B-instruct}; the released replay-cohort manifest maps the GPT \ragmode row to a different run. The open-source rows of Table~\ref{tab:replay} come from reruns that load the embedder and score both access modes in one run; every turn retrieves three sections, identical across the three readers and to the frontier heuristic-planner control on all 520 turns (\path{analysis/iclr/open_source_embed_rerun.json}). The submitted version reported earlier open-source \ragmode runs that requested the same MiniLM embedder but loaded none, so the embedding score was zero for every scored section. Those runs retrieved with lexical, heading, tag, and year matching only, returned fewer than three sections on 43 of 520 turns and none on 9, and matched the heuristic control on 71 of 520 turns. Table~\ref{tab:may-open-source} keeps their rows as a configuration record: their \ragmode losses were 6.2, 8.1, and 9.6 points for Qwen, Llama, and Gemma, against 2.1, 2.3, and 6.5 in the reruns. The reruns also use an updated runner and the revised SF prompt, so these judged losses mix the embedder with a change in scoring; Table~\ref{tab:os-key} compares the two versions under SF$_\text{key}$. An attempted repeat of the Sonnet \ragmode run requested \texttt{Alibaba-NLP/gte-Qwen2-7B-instruct}, loaded no embedder, and retrieved no section on 54 of 520 turns; it is excluded from all reported analyses. Source artifacts: \path{analysis/iclr/retrieval_embedder_audit.json}, \path{traces/t3_gpt_retrieve.jsonl}, \path{traces/t3_sonnet_retrieve.jsonl}, and \path{data/replay_cohort/manifest.json}. Table~\ref{tab:config} records the configuration of every generation, replay, adaptation, and retrieval component, and Table~\ref{tab:history} gives the composition of the scripted histories.

\begin{table}[htbp]
\caption{Recorded generation, replay, adaptation, and retrieval configuration. Provider aliases and historical local jobs did not retain resolved API revisions, model commit hashes, or a complete package lockfile; these missing pins are disclosed and not reconstructed from a later environment.}
\label{tab:config}
\centering
\small
\setlength{\tabcolsep}{3pt}
\renewcommand{\arraystretch}{1.1}
\begin{tabularx}{\textwidth}{@{}>{\raggedright\arraybackslash}p{0.15\textwidth}>{\raggedright\arraybackslash}p{0.34\textwidth}Y@{}}
\toprule
Component & Model identifier & Decoding and retrieval \\
\midrule
GPT generation & \texttt{gpt-5.1}; provider alias, revision not logged & Minimal reasoning; provider-default sampling; stage-specific output caps. \\
GPT-mini generation & \texttt{gpt-5-mini}; provider alias, revision not logged & Minimal reasoning; provider-default sampling; stage-specific output caps. \\
Sonnet generation & \texttt{claude-sonnet-4-6} & Temperature 0 for structured calls and 0.2 for prose; stage-specific caps. \\
Haiku generation & \texttt{claude-haiku-4-5-20251001} & Temperature 0 for structured calls and 0.2 for prose; stage-specific caps. \\
Qwen generation & \path{Qwen/Qwen3-32B}; commit not logged & bf16; $T=0.2$; top-$p=0.9$; repetition penalty 1.02; at most 1,400 new tokens. \\
Llama generation & \path{meta-llama/Llama-3.1-70B-Instruct}; commit not logged & bf16; $T=0.2$; top-$p=0.9$; repetition penalty 1.02; at most 1,600 new tokens. \\
Gemma generation & \path{google/gemma-4-26B-A4B-it}; commit not logged & bf16; $T=0.2$; top-$p=0.9$; repetition penalty 1.02; at most 1,200 new tokens. \\
Frontier replay & \texttt{gpt-5.1}; \texttt{claude-sonnet-4-6} & GPT minimal reasoning; Sonnet $T=0$; at most 120 output tokens; target-model query planner. \\
Open-source replay & Generation identifiers above & bf16 greedy decoding; repetition penalty 1.02; at most 220 output tokens; heuristic query planner; embedder loaded; both access modes in one run. \\
Retrieval and scoring & \texttt{all-MiniLM-L6-v2}; \texttt{gpt-5.1} judge & Normalized embeddings; hybrid section retrieval; $k=3$; score floor 0.08; 550-character excerpts; two-turn query history. \\
Gemma depth runs & Open-source replay model & Heuristic planner; $k\in\{3,12\}$ under a 1,650-character total; score floor 0.08; at most 220 output and judge tokens; memoized judge calls. \\
Gemma adaptation & QLoRA adapter on the Gemma replay model; state chains annotated by \texttt{claude-sonnet-4-6} & 4-bit base; rank 16, $\alpha=32$, dropout 0.05 on attention projections; learning rate $5\times10^{-5}$; one epoch; effective batch 8; at most 3,072 tokens; seed 13. \\
Frontier depth runs & \texttt{gpt-5.1}; \texttt{claude-sonnet-4-6}; \texttt{gpt-5.1} judge & Heuristic planner; $k\in\{3,12\}$ under a 1,650-character total; score floor 0.08; at most 220 output and judge tokens; memoized judge calls. \\
\bottomrule
\end{tabularx}
\end{table}

\begin{table}[htbp]
\caption{Open-source rows of the submitted version, kept as a configuration record. These \ragmode runs loaded no embedder; Table~\ref{tab:replay} replaces them with embedder-loaded reruns. All-turn pass rates (\%) and persona-bootstrap 95\% intervals for \ragmode$-$\docmode.}
\label{tab:may-open-source}
\centering
\small
\begin{tabular}{llrrrrr}
\toprule
Access & Target & Overall & SF & IF & NIP & \ragmode$-$\docmode \\
\midrule
\docmode & Qwen & 38.46 & 46.94 & 19.80 & 18.64 & \\
\ragmode & Qwen & 32.31 & 34.44 & 33.66 & 16.95 & $-6.2$ [$-13.7$, $+1.5$] \\
\docmode & Llama & 70.38 & 87.22 & 39.60 & 20.34 & \\
\ragmode & Llama & 62.31 & 73.06 & 37.62 & 38.98 & $-8.1$ [$-14.0$, $-2.5$] \\
\docmode & Gemma & 80.77 & 82.22 & 77.23 & 77.97 & \\
\ragmode & Gemma & 71.15 & 68.06 & 79.21 & 76.27 & $-9.6$ [$-17.9$, $-1.3$] \\
\bottomrule
\end{tabular}
\end{table}

\begin{table}[htbp]
\caption{Conversation-history composition of the frozen replay slice. Retention episodes probe supported facts, routines, and artifacts; time-jump episodes require phase selection; adversarial episodes test false-premise rejection and unsupported extrapolation. Prefilled turns are scripted history; live turns are rated target-model responses.}
\label{tab:history}
\centering
\small
\begin{tabular}{lrrrr}
\toprule
Episode type & Episodes & Prefilled & Live & Prefilled (\%) \\
\midrule
Retention & 32 & 2,864 & 256 & 91.8 \\
Time jump & 32 & 2,744 & 136 & 95.3 \\
Adversarial & 32 & 2,752 & 128 & 95.6 \\
\midrule
All & 96 & 8,360 & 520 & 94.1 \\
\bottomrule
\end{tabular}
\end{table}

\section{Replay Uncertainty and Answer Availability}
\label{app:replay-uncertainty}

\begin{figure}[htbp]
\centering
\includegraphics[width=0.8\textwidth]{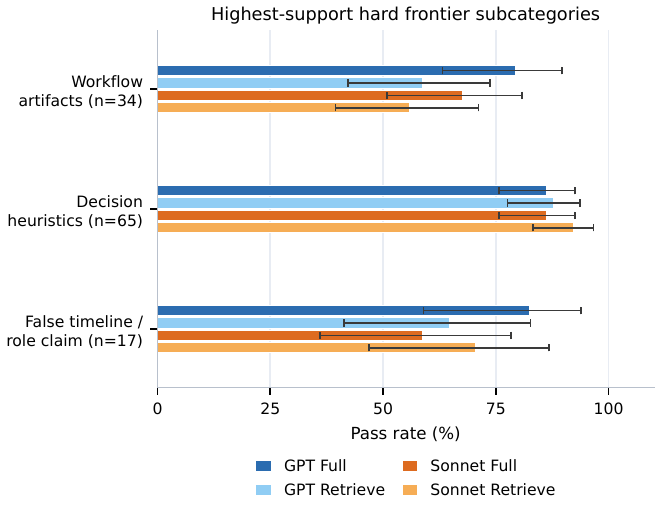}
\caption{Frontier pass rates for the highest-support hard subcategories, with Wilson 95\% intervals; support counts are rated turns. Workflow-artifact items fall under \ragmode for both targets, while decision-heuristic items stay above 86\% in every condition.}
\label{fig:hard-subcats}
\end{figure}

Table~\ref{tab:replay-full} gives the counts and Wilson intervals behind Table~\ref{tab:replay}, and Table~\ref{tab:replay-boot} gives persona-clustered intervals for each row and each paired access contrast. Table~\ref{tab:availability} separates answer availability from answer quality for the two open-source targets affected by output extraction. Table~\ref{tab:frontier-bucket} breaks frontier replay down by dossier length for the runs of Table~\ref{tab:replay} and for two packagings of heuristic-planner retrieval, which Section~\ref{sec:extrinsic} compares. Figure~\ref{fig:hard-subcats} reports the frequent subcategories with the lowest frontier pass rates.

\begin{table}[htbp]
\caption{Full replay counts behind Table~\ref{tab:replay}. Each cell gives passed/rated turns, the all-turn pass rate, and the Wilson 95\% half-width, $(\mathrm{hi}-\mathrm{lo})/2$, in percentage points. Nonanswers fail.}
\label{tab:replay-full}
\centering
\small
\setlength{\tabcolsep}{1.5pt}
\begin{tabular}{llllll}
\toprule
Access & Target & Overall & SF & IF & NIP \\
\midrule
\docmode & GPT & 479/520 (92.12 $\pm$2.3) & 339/360 (94.17 $\pm$2.5) & 88/101 (87.13 $\pm$6.6) & 52/59 (88.14 $\pm$8.3) \\
\docmode & Sonnet & 461/520 (88.65 $\pm$2.7) & 331/360 (91.94 $\pm$2.8) & 90/101 (89.11 $\pm$6.1) & 40/59 (67.80 $\pm$11.6) \\
\ragmode & GPT & 434/520 (83.46 $\pm$3.2) & 291/360 (80.83 $\pm$4.1) & 92/101 (91.09 $\pm$5.7) & 51/59 (86.44 $\pm$8.8) \\
\ragmode & Sonnet & 437/520 (84.04 $\pm$3.1) & 301/360 (83.61 $\pm$3.8) & 90/101 (89.11 $\pm$6.1) & 46/59 (77.97 $\pm$10.4) \\
\docmode & Qwen & 195/520 (37.50 $\pm$4.1) & 164/360 (45.56 $\pm$5.1) & 19/101 (18.81 $\pm$7.6) & 12/59 (20.34 $\pm$10.1) \\
\ragmode & Qwen & 184/520 (35.38 $\pm$4.1) & 144/360 (40.00 $\pm$5.0) & 31/101 (30.69 $\pm$8.9) & 9/59 (15.25 $\pm$9.1) \\
\docmode & Llama & 392/520 (75.38 $\pm$3.7) & 334/360 (92.78 $\pm$2.7) & 41/101 (40.59 $\pm$9.4) & 17/59 (28.81 $\pm$11.3) \\
\ragmode & Llama & 380/520 (73.08 $\pm$3.8) & 316/360 (87.78 $\pm$3.4) & 35/101 (34.65 $\pm$9.1) & 29/59 (49.15 $\pm$12.4) \\
\docmode & Gemma & 455/520 (87.50 $\pm$2.8) & 315/360 (87.50 $\pm$3.4) & 83/101 (82.18 $\pm$7.4) & 57/59 (96.61 $\pm$5.3) \\
\ragmode & Gemma & 421/520 (80.96 $\pm$3.4) & 290/360 (80.56 $\pm$4.1) & 74/101 (73.27 $\pm$8.5) & 57/59 (96.61 $\pm$5.3) \\
\bottomrule
\end{tabular}
\end{table}

\begin{table}[htbp]
\caption{Length-stratified persona-clustered bootstrap intervals (95\%) for all-turn replay rates and paired access-mode differences, in percent and percentage points. Each of 10,000 resamples draws personas with replacement within the four length strata and keeps all 520 paired turns of each drawn persona. Qwen and Gemma use the post-extraction scores of Table~\ref{tab:replay}.}
\label{tab:replay-boot}
\centering
\small
\begin{tabular}{lrrr}
\toprule
Target & \docmode & \ragmode & \ragmode$-$\docmode \\
\midrule
GPT-5.1 & 92.1 [89.6, 94.6] & 83.5 [78.5, 88.3] & $-8.7$ [$-13.7$, $-3.8$] \\
Sonnet 4.6 & 88.7 [85.8, 91.2] & 84.0 [79.4, 88.5] & $-4.6$ [$-10.0$, $+0.6$] \\
Qwen & 37.5 [31.0, 44.0] & 35.4 [28.5, 42.5] & $-2.1$ [$-9.6$, $+5.4$] \\
Llama & 75.4 [71.2, 79.6] & 73.1 [69.2, 76.7] & $-2.3$ [$-6.7$, $+2.1$] \\
Gemma & 87.5 [81.9, 92.1] & 81.0 [76.5, 84.8] & $-6.5$ [$-13.8$, $+0.8$] \\
\bottomrule
\end{tabular}
\end{table}

\begin{table}[htbp]
\caption{Answer availability for the open-source targets affected by output extraction, in percent of 520 turns. Qwen exposes an eligible answer on 272 \docmode and 284 \ragmode turns. Every Gemma turn has an eligible leading answer, although trailing control text follows it on 503 \docmode and 520 \ragmode turns. GPT, Sonnet, and Llama return an eligible answer on every turn.}
\label{tab:availability}
\centering
\small
\begin{tabular}{llrrr}
\toprule
Access & Target & All-turn rate & Rate given answer & No eligible answer \\
\midrule
\docmode & Qwen & 37.50 & 71.69 & 47.69 \\
\ragmode & Qwen & 35.38 & 64.79 & 45.38 \\
\docmode & Gemma & 87.50 & 87.50 & 0.00 \\
\ragmode & Gemma & 80.96 & 80.96 & 0.00 \\
\bottomrule
\end{tabular}
\end{table}

\begin{table}[htbp]
\caption{Frontier replay by target-length bucket. Each cell gives passed turns out of the column's $n$ rated turns, the pass rate, and the Wilson 95\% half-width. The upper rows are the runs of Table~\ref{tab:replay}. The lower rows use the heuristic planner and retrieve identical sections at $k=3$; they differ in excerpt construction and in the answer-token limit. Every row is a separate run with separate judge calls.}
\label{tab:frontier-bucket}
\centering
\small
\setlength{\tabcolsep}{1.5pt}
\begin{tabular}{llllll}
\toprule
Access & Target & 1k ($n{=}136$) & 2.5k ($n{=}152$) & 5k ($n{=}128$) & 10k ($n{=}104$) \\
\midrule
\docmode & GPT & 134 (98.5 $\pm$2.4) & 133 (87.5 $\pm$5.3) & 114 (89.1 $\pm$5.4) & 98 (94.2 $\pm$4.7) \\
\docmode & Sonnet & 130 (95.6 $\pm$3.6) & 126 (82.9 $\pm$6.0) & 110 (85.9 $\pm$6.0) & 95 (91.3 $\pm$5.5) \\
\ragmode & GPT & 131 (96.3 $\pm$3.4) & 126 (82.9 $\pm$6.0) & 104 (81.2 $\pm$6.7) & 73 (70.2 $\pm$8.7) \\
\ragmode & Sonnet & 134 (98.5 $\pm$2.4) & 134 (88.2 $\pm$5.2) & 102 (79.7 $\pm$6.9) & 67 (64.4 $\pm$9.1) \\
\midrule
\multicolumn{6}{@{}l}{\ragmode, heuristic planner, $3\times450$ characters, 120 answer tokens} \\
& GPT & 125 (91.9 $\pm$4.7) & 120 (78.9 $\pm$6.4) & 106 (82.8 $\pm$6.5) & 70 (67.3 $\pm$8.9) \\
& Sonnet & 126 (92.6 $\pm$4.5) & 116 (76.3 $\pm$6.7) & 105 (82.0 $\pm$6.6) & 66 (63.5 $\pm$9.1) \\
\multicolumn{6}{@{}l}{\ragmode, heuristic planner, $k=3$, 1,650-character total, 220 answer tokens} \\
& GPT & 131 (96.3 $\pm$3.4) & 130 (85.5 $\pm$5.6) & 113 (88.3 $\pm$5.6) & 87 (83.7 $\pm$7.1) \\
& Sonnet & 131 (96.3 $\pm$3.4) & 121 (79.6 $\pm$6.4) & 113 (88.3 $\pm$5.6) & 89 (85.6 $\pm$6.8) \\
\bottomrule
\end{tabular}
\end{table}

\section{Supplementary Replay Diagnostics}
\label{app:supp-replay}

\paragraph{Temporal split.}
Table~\ref{tab:robustness} summarizes length retention and the temporal gap for the frontier readers, and Table~\ref{tab:temporal-sf} gives the SF pass rates behind the temporal gap.

\begin{table}[htbp]
\caption{Extrinsic robustness of frontier readers. Retention divides \ragmode by \docmode pass rates per length bucket (Appendix Table~\ref{tab:frontier-bucket}). The temporal gap is phase-linked minus timeless SF accuracy over 115 and 245 items, a post-hoc split (Appendix Table~\ref{tab:temporal-sf}).}
\label{tab:robustness}
\centering
\small
\setlength{\tabcolsep}{4pt}
\begin{tabular}{lrrrrrr}
\toprule
& \multicolumn{4}{c}{Length retention (\ragmode/\docmode)} & \multicolumn{2}{c}{Temporal gap (pts)} \\
\cmidrule(lr){2-5}\cmidrule(lr){6-7}
Reader & 1k & 2.5k & 5k & 10k & \docmode & \ragmode \\
\midrule
GPT & 0.978 & 0.947 & 0.912 & 0.745 & $+2.2$ & $-12.7$ \\
Sonnet & 1.031 & 1.063 & 0.927 & 0.705 & $-3.5$ & $-11.7$ \\
\bottomrule
\end{tabular}
\end{table}

\begin{table}[htbp]
\caption{SF pass rates (\%) over 115 phase-linked and 245 timeless items. $\Delta$ is phase-linked minus timeless in percentage points. This split was defined after the main replay analysis.}
\label{tab:temporal-sf}
\centering
\small
\setlength{\tabcolsep}{2.5pt}
\begin{tabular}{lrrrrrr}
\toprule
& \multicolumn{3}{c}{\docmode} & \multicolumn{3}{c}{\ragmode} \\
\cmidrule(lr){2-4}\cmidrule(lr){5-7}
Target & Phase & Timeless & $\Delta$ & Phase & Timeless & $\Delta$ \\
\midrule
GPT & 95.7 & 93.5 & $+2.2$ & 72.2 & 84.9 & $-12.7$ \\
Sonnet & 89.6 & 93.1 & $-3.5$ & 75.7 & 87.3 & $-11.7$ \\
\bottomrule
\end{tabular}
\end{table}

\paragraph{Qualitative failure patterns.} Three patterns recurred in a qualitative review of failed turns. A \emph{cross-phase application} uses an obsolete rule despite identifying the right topic. In \emph{false-premise compliance}, the answer accepts an incorrect date, role, or affiliation that the evidence contradicts. A \emph{fragment integration failure} follows one retrieved passage and overlooks a constraint found in another. Topical retrieval can therefore succeed in all three cases while the final answer still fails.

\paragraph{Encoding format.}
Table~\ref{tab:formats} gives the format contrasts summarized in Section~\ref{sec:item-family}. On the 32-persona panel, the three packet rows use identical top-three sections on all 520 core turns; on the screen, both packet types select the same records.

\begin{table}[htbp]
\caption{Recoverability by encoding format. Rows within a panel share items and retrieved evidence; the 32-persona rows are the Gemma runs of Table~\ref{tab:replay} and Section~\ref{sec:item-family}, scored with the extraction rule of Table~\ref{tab:replay}; the 8-persona rows were rescored with the same rule during discussion (Appendix~\ref{app:interventions} gives full-response values). IF, SF, and NIP totals are 101, 360, and 59 (upper panel) and 29, 109, and 16 (lower panel).}
\label{tab:formats}
\centering
\small
\setlength{\tabcolsep}{4pt}
\begin{tabular}{llrrrr}
\toprule
Panel & Encoding (Gemma, untuned) & IF & SF & NIP & Overall \\
\midrule
32 personas & Full dossier (\docmode) & 83 & 315 & 57 & 455 \\
(520 turns) & Raw sections, 3 $\times$ 550 characters & 74 & 290 & 57 & 421 \\
& Raw sections, 1,650-character total & 83 & 288 & 57 & 428 \\
& Versioned state records, same sections & 67 & 284 & 49 & 400 \\
\midrule
8 personas & Versioned state records, 1,650 / 8,000 & 21 / 16 & 64 / 88 & 14 / 11 & 99 / 115 \\
(154 turns) & Resolved-state packet, 1,650 / 8,000 & 20 / 22 & 75 / 88 & 12 / 10 & 107 / 120 \\
& Gold sections, 8,000 & 21 & 92 & 10 & 123 \\
\bottomrule
\end{tabular}
\end{table}

\paragraph{Retrieval diagnostics.}
Frontier retrieval reaches the part of the dossier that governs each item. Of the 208 unique frontier items, 174 carry phase labels and 188 carry section labels. Counting each item's first rated occurrence, every phase-labeled item has a phase hit@3, and section hit@3 covers 178 (GPT) and 183 (Sonnet) of the 188 section-labeled items; over the 520 rated turns, phase-pure retrieval is 94.4\% and 93.1\%. Source artifact: \path{analysis/audit_and_replay/retrieval_phase_hits_publication_rag.json}.

\paragraph{Planner substitution.}
Frontier \ragmode results do not require target-authored queries. A fixed heuristic planner keeps phase hit@3 at 100\% but lowers section retrieval; GPT moves from 434 to 421 of 520 turns (planned-minus-heuristic $+2.5$ points, 95\% CI $[-2.3,+7.4]$) and Sonnet from 437 to 413 ($+4.6$ points, 95\% CI $[-0.3,+9.0]$). Both intervals include zero; no planner effect is established. The heuristic control also used 450-character excerpts and a 0.18 score floor, so the contrast changes excerpt length together with the planner.

\paragraph{Failure attribution by reader.}
Table~\ref{tab:attribution} applies the strict key rule to the two frontier \ragmode rows and the depth contrast. GPT-5.1, Sonnet~4.6, and Gemma receive byte-identical packets on all 520 turns at each depth; Gemma uses the post-extraction decisions of Section~\ref{sec:item-family}. At $k=3$, pass rates with packet key support are 98.2\%, 98.2\%, and 94.9\%; without it they are 71.3\%, 70.6\%, and 57.3\%. At $k=12$, the strict key rule assigns zero retrieval failures and more packaging failures to every reader. Source artifacts: \path{analysis/iclr/attribution_cleanup_audit.json} and \path{analysis/iclr/attribution_strict_ledgers.jsonl}. The earlier \path{analysis/iclr/sf_packet_attribution_extended.json} records the submitted loose rule and mixed Gemma scoring for historical comparison.

\begin{table}[htbp]
\caption{Strict key-based attribution on 360 SF turns. Frozen scored misses are assigned to answer formation when the packet satisfies the key, source support (``No key'') when all labeled sections fail the key, packaging when the selected labeled-section union satisfies it but the packet fails it, and retrieval otherwise. Gemma uses post-extraction scoring. The last column counts passes without detected packet support.}
\label{tab:attribution}
\centering
\small
\setlength{\tabcolsep}{3.5pt}
\begin{tabular}{llrrrrrrr}
\toprule
Evidence & Reader & Pass & Fail & Retr. & Pack. & Answer & No key & Pass, no support \\
\midrule
Table~\ref{tab:replay} \ragmode & GPT-5.1 & 291 & 69 & 11 & 21 & 23 & 14 & 30 \\
(model planner, $3\times550$) & Sonnet 4.6 & 301 & 59 & 8 & 26 & 16 & 9 & 28 \\
\midrule
Identical packets & GPT-5.1 & 315 & 45 & 11 & 19 & 4 & 11 & 102 \\
($k=3$, 1,650 total) & Sonnet 4.6 & 314 & 46 & 10 & 20 & 4 & 12 & 101 \\
& Gemma & 288 & 72 & 17 & 33 & 11 & 11 & 82 \\
\midrule
Identical packets & GPT-5.1 & 312 & 48 & 0 & 40 & 2 & 6 & 87 \\
($k=12$, 1,650 total) & Sonnet 4.6 & 302 & 58 & 0 & 50 & 6 & 2 & 81 \\
& Gemma & 282 & 78 & 0 & 51 & 18 & 9 & 73 \\
\bottomrule
\end{tabular}
\end{table}

Table~\ref{tab:context-check} checks the passes without detected packet support against the other context each reader saw: the compact profile in the system prompt and the earlier turns of the episode.

\begin{table}[htbp]
\caption{Context check for SF attribution. Passes without packet support are split by key support in the profile, earlier turns only, or neither. ``Judged'' is a subset of ``Neither'' and describes a scoring route. The final columns use the strict packaging counts: key absent from profile and history, and item-check flags (Appendix~\ref{app:item-check}). Availability does not establish source use.}
\label{tab:context-check}
\centering
\small
\setlength{\tabcolsep}{3pt}
\begin{tabular}{llrrrrrrr}
\toprule
Evidence & Reader & No support & Profile & Earlier & Neither & Judged & Key absent & Flagged \\
\midrule
Table~\ref{tab:replay} \ragmode & GPT-5.1 & 30 & 8 & 13 & 9 & 3 & 19 of 21 & 6 of 21 \\
& Sonnet 4.6 & 28 & 8 & 13 & 7 & 3 & 24 of 26 & 8 of 26 \\
$k=3$, 1,650 total & GPT-5.1 & 102 & 48 & 32 & 22 & 15 & & \\
& Sonnet 4.6 & 101 & 50 & 30 & 21 & 15 & & \\
\bottomrule
\end{tabular}
\end{table}

\paragraph{Aligned key-free support audit.}
The 128 frontier SF misses comprise 27 unique questions from 20 personas. Both the deterministic rule and the key-free judges receive the same realized packet, selected labeled-section union, and all-labeled-section union, using canonical body text and the same decision order. The earlier key-free check rated sections individually; 9 of the 19 strict-key retrieval cases require joined sections under the matcher. GPT-5.1 and Sonnet~4.6 retain the earlier question-only support rubric and answer with keys and reader outputs withheld. Exact cached requests are reused. Table~\ref{tab:aligned-attribution} reports each measure separately. Across strict packaging cases, the two judges retain the label on 17 of 21 GPT and 21 of 26 Sonnet turns. These are model-based support measurements.

\begin{table}[htbp]
\caption{Attribution under aligned evidence units. Key rows use deterministic support; judge rows count labels assigned by both model judges. ``Disagree'' retains different judge labels. All rows concern the same frozen scored misses.}
\label{tab:aligned-attribution}
\centering
\small
\begin{tabular}{llrrrrr}
\toprule
Reader & Support measure & Retr. & Pack. & Answer & Source & Disagree \\
\midrule
GPT-5.1 & Key & 11 & 21 & 23 & 14 & 0 \\
& Both judges & 7 & 19 & 25 & 9 & 9 \\
Sonnet 4.6 & Key & 8 & 26 & 16 & 9 & 0 \\
& Both judges & 4 & 22 & 18 & 7 & 8 \\
\bottomrule
\end{tabular}
\end{table}

The separate full-dossier check rates 24 of 27 questions fully answerable for GPT-5.1 and 26 for Sonnet, with 24 receiving that rating from both. It assesses question answerability; the earlier keyed screen assesses support for the specified answer. The 60 seeded pass controls were selected for deterministic packet support. GPT-5.1 and Sonnet rate 24/30 and 23/30 GPT-reader packets fully supporting, and 25/30 and 27/30 Sonnet-reader packets; scored passes do not establish evidence sufficiency. Human raters were unavailable for this audit; its question, source, and packet checks remain model-based. Stored scores are preserved, and key-positive failed responses and support elsewhere in the profile or history are recorded as audit flags. Source artifacts: \path{analysis/iclr/attribution_aligned_cleanup.json}, \path{analysis/iclr/attribution_aligned_ledger.jsonl}, and \path{analysis/iclr/attribution_source_audit.json}.

\begin{table}[htbp]
\caption{Whole-section outcomes by the strict label in the 550-character matched control, scored by the answer key. Both policies share scripted episodes and selected sections, with live answers generated separately. The supported-item row restricts to keys both original validators rated supported.}
\label{tab:attribution-repair}
\centering
\small
\begin{tabular}{lrr}
\toprule
Label in excerpt condition & GPT-5.1 & Sonnet 4.6 \\
\midrule
Packaging & 29/32 & 25/30 \\
Packaging, supported items & 21/23 & 17/21 \\
Answer formation & 2/8 & 1/8 \\
\bottomrule
\end{tabular}
\end{table}

\paragraph{Gold-section and fixed-budget design.}
Items without section labels, mainly prompt-injection refusals and card-grounded items, retain the compact card in the gold-section condition and are reported separately; the condition controls evidence selection but retains the bounded packet and downstream answer process. The retrieval-budget study holds the query planner, answer prompt, conversation history, target model, and scorer fixed while varying $k\in\{1,3,5\}$. Under the \emph{per-passage} policy, each passage receives up to 550 characters, and evidence volume grows with $k$; under the \emph{fixed-total} policy, at most 1,650 characters are distributed across selected passages. The gold-section condition uses the same 1,650-character cap. Every turn records selected section IDs, gold hit and recall, realized evidence characters, per-section allocation, the effective dense embedder, tokens, and cost. The fixed-total diagnostic uses eight personas, two from each length bucket, and 130 rated turns per condition; it tests the intervention mechanics and does not replace the 32-persona, length-stratified comparison.

\paragraph{Fixed-budget evidence diagnostic.}
Perfect section selection adds little on this panel: under the same packaging, gold sections add zero net passes over $k=3$ for GPT-5.1 and three for Sonnet~4.6 (Table~\ref{tab:budget-diagnostic}). Both targets pass over 94\% of turns with gold sections, leaving little room for a retrieval gain on this panel. The counts bound the observed return to retrieval improvement; they do not measure answer quality.

\begin{table}[htbp]
\caption{Eight-persona fixed-total diagnostic: answer passes and mean labeled-section recall.}
\label{tab:budget-diagnostic}
\centering
\small
\setlength{\tabcolsep}{4pt}
\begin{tabular}{lrrrrrr}
\toprule
Target & Gold & $k=1$ & $k=3$ & $k=5$ & Recall@$3$ & Recall@$5$ \\
\midrule
GPT-5.1 & 123/130 & 115/130 & 123/130 & 122/130 & 76.2\% & 88.2\% \\
Sonnet 4.6 & 124/130 & 111/130 & 121/130 & 121/130 & 80.1\% & 89.3\% \\
\bottomrule
\end{tabular}
\end{table}

Top-$3$ scores highest on this panel. Moving from $k=1$ to $k=3$ adds 8 GPT and 10 Sonnet passes (paired wins/losses 11/3 and 11/1). Moving to $k=5$ raises mean labeled-section recall by 12.0 and 9.2 points without an answer gain, consistent with dilution under the fixed total: across 93 SF turns, labeled-section selections rise slightly (88 to 91 for GPT, 91 to 93 for Sonnet) while key-satisfying packets fall (80 to 76 and 80 to 77). These are not population estimates, and the packet audit covers SF only; IF and NIP lack sufficiency labels. Source artifacts: \path{analysis/iclr/experiment_summary_gpt51.json}, \path{analysis/iclr/experiment_summary_sonnet46.json}, and the \path{traces/diagnostic_*} files.

\section{Prompts, Rubrics, Exclusions, and Released Tables}

Replay prompt construction, query planning, evidence selection, fixed-total allocation, response extraction, and scoring are implemented in \path{code/replay/eval_episode_runner.py}. The gold-section and retrieval-budget conditions are defined in \path{code/replay/run_iclr_retrieval_experiments.py}, which also keeps the earlier weighted metadata matcher for replication; aggregate comparisons are produced by \path{code/replay/analyze_iclr_retrieval_experiments.py}. The submitted SF packet attribution was produced by \path{code/replay/analyze_sf_packet_attribution.py}. Corrected strict and aligned analyses are documented by \path{code/analysis/iclr/build_attribution_cleanup_audit.py} and \path{code/analysis/iclr/attribution_aligned_cleanup.py}; \path{code/analysis/iclr/recompute_attribution.py} verifies the released ledgers. Phase-evaluation rubrics and human instructions are indexed by \path{data/cited_artifacts/artifact_manifest.md}. Redacted turn-level traces for every reported replay condition are in \path{traces/}, and \path{code/analysis/iclr/recompute_from_traces.py} recomputes the reported replay numbers from them.

The replay-cohort README records the blinded exclusions and explains why 431 frozen definitions produce 520 rated turns. Per-row episode aggregates and analysis outputs retain family counts, clustered uncertainty, temporal strata, answer availability, judge substitution, retrieval diagnostics, and SF packet attribution. Source artifacts, under \path{analysis/audit_and_replay/}: \path{replay_clustered_uncertainty_summary.json}, \path{replay_temporal_stratified_summary.json}, \path{open_source_output_sensitivity_summary.json}, \path{judge_sensitivity_summary.json}, \path{retrieval_phase_hits_publication_rag.json}, and \path{sf_packet_attribution_summary.json}; see also \path{data/replay_cohort/README.md}.

\section{Intervention Diagnostics}
\label{app:interventions}

\paragraph{Gold-section evidence.}
The gold-section condition builds each packet from the exact labeled section identifiers, after a check that every identifier resolves to a stored section, and enforces the same total evidence cap as the fixed-budget \ragmode conditions. The eight-persona diagnostic covers both frontier targets, with 130 rated turns per condition. Gold-section evidence is evaluated only on eight-persona panels: this diagnostic and the 8,000-character screen of the encoding-format comparison (Appendix~\ref{app:supp-replay}). Source artifacts: \path{code/replay/run_iclr_retrieval_experiments.py}, \path{code/replay/eval_episode_runner.py}, and the \path{traces/diagnostic_*} files.

\paragraph{Output delimitation.}
Gemma usually continues past its end-of-turn marker. Tables~\ref{tab:replay}, \ref{tab:qtypes}, \ref{tab:item-family}, and~\ref{tab:formats} (upper panel) and Section~\ref{sec:item-family} score only the text before the marker. Scoring the full response instead, as the submitted version did for Section~\ref{sec:item-family}, gives a depth IF gain of 14.9 points ($p=0.028$), an adaptation IF loss of 21.8 ($p=0.0001$) and NIP gain of 11.9 ($p=0.039$), an encoding NIP loss of 11.9 ($p=0.016$), and an encoding overall change of 6 turns ($p=0.58$). The eight-persona screen conditions were rescored with the extraction rule during discussion (Table~\ref{tab:formats}, lower panel). On the full response, the resolved-state packet passed 122 turns against 111 for versioned records at 8,000 characters ($p=0.019$), and the ten adapters passed 101 to 110 turns against 108 and 104 for two gradings of the untuned model. Source artifacts: \path{analysis/iclr/gemma_interventions_postextraction.json} and \path{analysis/iclr/rebuttal_screen_contrasts.json}.

\paragraph{Parameter-efficient adaptation diagnostics.}
The earlier adaptation stages show reliable format and citation-control effects and no answer-accuracy gain; the main adaptation contrast in Section~\ref{sec:item-family} uses the SFT adapter from these experiments. An initial extractive QLoRA pilot on Gemma~4 changed matched replay from 102/130 to 105/130 (7 adapted-only and 4 base-only passes), and under the same chain evidence, a later state-chain adapter passed 134/154 against 129/154 for the SFT adapter; both changes are inconclusive by exact McNemar ($p=0.549$ and $p=0.227$). The state-chain adapter raised exact output-schema compliance from 30/154 to 151/154. Sparse-active refinement then reduced overcitation from 79/128 to 45/128 defined-target turns while overall answer accuracy stayed at 104/154 and minimal-support satisfaction changed from 109/128 to 108/128. Across four correction-stage seeds, per-turn citation precision rose from 69.8\% untuned to 74.0--76.8\%. Counts in this paragraph score the full response. Source artifacts: \path{analysis/iclr/sparse_active_screen_analysis.json}, \path{analysis/iclr/seed_noise_floor_analysis.json}, and the \path{traces/screen_*} files; the analyses of the extractive and state-chain pilots are not in the release.

\section{Controls Added During Discussion}
\label{app:rebuttal}

\paragraph{Length and access.} Each persona is written at one length. Pairing \docmode and \ragmode estimates access differences within each persona; length comparisons remain confounded with persona and item composition. Table~\ref{tab:length-interaction} gives the change in the \ragmode-minus-\docmode gap from 1k to 10k under several restrictions and scorings. Source artifacts: \path{analysis/iclr/rebuttal_tier0.json} and \path{analysis/iclr/rebuttal_cap_interaction.json}.

\begin{table}[htbp]
\caption{Change from 1k to 10k words in the \ragmode$-$\docmode gap (points), with length-stratified persona-bootstrap 95\% intervals (10,000 resamples). Standardization reweights each bucket to the pooled item-family and subcategory mix. The Sonnet-judge row replaces every GPT-5.1 judgment with the Sonnet 4.6 judgment. The regenerated row uses the answers of Table~\ref{tab:cap-rerun}, and the row below it re-judges the original answers in the same run. SF$_\text{key}$ rows score SF turns by the answer key alone; the rerun row repeats the Table~\ref{tab:replay} \ragmode setting with the current runner. The last row drops SF items flagged by the item check (Appendix~\ref{app:item-check}).}
\label{tab:length-interaction}
\centering
\small
\begin{tabular}{lrr}
\toprule
Analysis & GPT-5.1 & Sonnet 4.6 \\
\midrule
All turns & $-21.8$ [$-40.8$, $-3.3$] & $-29.9$ [$-47.7$, $-12.0$] \\
\docmode change alone & $-4.3$ [$-9.1$, $+0.3$] & $-4.2$ [$-9.3$, $+1.6$] \\
Family and subcategory standardized & $-20.1$ [$-39.5$, $-2.9$] & $-19.4$ [$-44.4$, $-4.2$] \\
SF turns only & $-31.4$ [$-53.8$, $-7.9$] & $-36.3$ [$-62.7$, $-11.1$] \\
SF identity and context only & $-35.9$ [$-60.9$, $-9.4$] & $-40.6$ [$-68.8$, $-12.5$] \\
Sonnet 4.6 judge & $-24.5$ [$-43.3$, $-6.2$] & $-27.1$ [$-46.9$, $-9.3$] \\
Truncated answers regenerated & $-22.8$ [$-42.0$, $-3.6$] & $-29.9$ [$-48.1$, $-12.5$] \\
Same answers re-judged & $-19.9$ [$-37.4$, $-2.8$] & $-29.9$ [$-46.9$, $-13.3$] \\
SF$_\text{key}$, SF turns & $-24.0$ [$-48.3$, $-2.0$] & $-18.0$ [$-36.0$, $-2.0$] \\
SF$_\text{key}$, \ragmode setting rerun & $-22.3$ [$-43.8$, $-2.9$] & $-16.0$ [$-34.8$, $+1.0$] \\
SF turns, flagged items removed & $-28.8$ [$-51.6$, $-7.0$] & $-29.0$ [$-52.6$, $-6.1$] \\
\bottomrule
\end{tabular}
\end{table}

\paragraph{Full-dossier dip at 2.5k and 5k.} \docmode accuracy is lower at 2.5k and 5k than at 1k and 10k (Table~\ref{tab:frontier-bucket}). GPT's 19 \docmode failures at 2.5k fall on 6 of the 8 personas, at most 6 on one; Sonnet's 26 fall on all 8, at most 8 on one. Leaving out any one persona keeps the 2.5k \docmode rate between 85.7 and 90.2\% (GPT) and 81.2 and 86.5\% (Sonnet). At 2.5k, GPT passes 85 of 96 SF and 18 of 23 NIP turns under \docmode, against 100 of 101 SF at 1k; the 2.5k bucket holds 23 NIP turns against 8 at 1k and at 10k, and nine 2.5k turns fail under \docmode for both readers, against 1, 4, and 0 in the other buckets.

\paragraph{Excerpt coverage.} Table~\ref{tab:coverage} relates labeled-section length, the share of a selected labeled section shown in the packet, and \ragmode SF accuracy. In persona-clustered logistic models of \ragmode SF accuracy, the log-length coefficient is $-0.95$ ($p=0.002$) alone and $-0.27$ ($p=0.56$) with coverage for GPT, and $-1.49$ ($p=0.0002$) and $-0.58$ ($p=0.35$) for Sonnet. With packet key support in place of coverage, it is $-0.75$ ($p=0.054$) and $-1.61$ ($p=0.004$). Nearly every 1k turn shows at least half of its selected labeled section and no 5k or 10k turn does, so coverage and length are nearly collinear. Among turns shown less than half of the section, accuracy falls from 79\% to 70\% (GPT) and from 90\% to 72\% (Sonnet) between 2.5k and 10k. On turns whose packet satisfies the key, accuracy is 79.6\% and 76.4\% at 10k, against 89.0 to 100\% at shorter lengths, so part of the 10k loss arises after the answer reaches the packet.

\begin{table}[htbp]
\caption{\ragmode SF turns of Table~\ref{tab:replay} by length bucket (GPT / Sonnet). Coverage is the share of a selected labeled section shown in the packet, in characters (the largest over selected labeled sections; zero when none is selected). Key in packet: the frozen key is satisfied by the packet.}
\label{tab:coverage}
\centering
\small
\setlength{\tabcolsep}{4pt}
\begin{tabular}{lrrrrrr}
\toprule
Bucket & $n$ & Median section chars & Coverage & Key in packet (\%) & Pass (\%) & Pass given key (\%) \\
\midrule
1k & 101 & 579 & .83 / .85 & 94.1 / 95.0 & 97.0 / 100.0 & 100.0 / 100.0 \\
2.5k & 96 & 1,503 & .32 / .33 & 85.4 / 87.5 & 79.2 / 88.5 & 89.0 / 97.6 \\
5k & 88 & 2,490 & .16 / .16 & 60.2 / 61.4 & 78.4 / 78.4 & 94.3 / 98.1 \\
10k & 75 & 3,929 & .08 / .08 & 72.0 / 73.3 & 64.0 / 61.3 & 79.6 / 76.4 \\
\bottomrule
\end{tabular}
\end{table}

\paragraph{Answer length.} The runner retries truncated-looking IF and NIP answers but not SF answers. We regenerated every rated frontier answer of Table~\ref{tab:replay} flagged as potentially truncated after reaching the 120-token limit (101 turns) from its identical message history with a 1,200-token limit, and scored the original and the regenerated answer with the original judge. For GPT, the rebuilt histories reproduce the original input token counts on all 69 turns. Re-judging the unchanged answers raises their passes from 77 to 87, so part of the gain is judge variation (Table~\ref{tab:cap-rerun}). Against the re-judged answers, the regenerated ones move the \ragmode-minus-\docmode gaps from $-8.5$ to $-9.0$ (GPT) and leave Sonnet at $-4.0$, and the length interaction changes by $-2.9$ points for GPT ($[-5.8,-1.0]$) and $0.0$ for Sonnet. Source artifact: \path{analysis/iclr/rebuttal_cap_interaction.json}. Source artifacts: \path{analysis/iclr/rebuttal_cap_rerun.py} and its summary.

\begin{table}[htbp]
\caption{Regenerated truncated answers. Columns 3--5 count passes on the regenerated turns: original decision, the original answer re-judged, and the regenerated answer. Row and IF totals give the same three versions over all 520 turns.}
\label{tab:cap-rerun}
\centering
\small
\begin{tabular}{lrrrrrr}
\toprule
Row & Turns & Original & Re-judged & Regenerated & Row total & IF total \\
\midrule
GPT-5.1 \docmode & 32 & 23 & 25 & 31 & 479 / 481 / 487 & 88 / 90 / 96 \\
GPT-5.1 \ragmode & 37 & 30 & 33 & 36 & 434 / 437 / 440 & 92 / 95 / 98 \\
Sonnet 4.6 \docmode & 18 & 16 & 17 & 18 & 461 / 462 / 463 & 90 / 90 / 90 \\
Sonnet 4.6 \ragmode & 14 & 8 & 12 & 13 & 437 / 441 / 442 & 90 / 94 / 95 \\
\bottomrule
\end{tabular}
\end{table}

\paragraph{Matched packaging control.} Every variant retrieves the same three sections on all 520 core turns with the heuristic planner ($k=3$, score floor 0.08), a 220-token answer limit, and one memoized judge with the revised SF prompt; only the excerpt changes (Table~\ref{tab:packaging}). Whole sections pass more turns than 550-character excerpts: GPT $+29/-10$ (exact $p=0.003$; persona-bootstrap $+3.7$ points $[0.0,+8.3]$, sign-flip $p=0.10$) and Sonnet $+28/-17$ ($p=0.14$; $+2.1$ $[-1.0,+5.6]$, $p=0.31$). We detect no difference between 450- and 550-character excerpts. Under SF$_\text{key}$, whole sections minus 550-character excerpts change the length interaction by $+18.0$ for GPT ($[+3.1,+38.2]$) and $+12.0$ for Sonnet ($[-3.4,+34.4]$); the model-written query of Table~\ref{tab:replay} minus the 550-character keyword variant gives $+0.7$ ($[-6.7,+7.6]$) and $+1.1$ ($[-10.7,+14.1]$). Source artifacts: \path{analysis/iclr/rebuttal_packaging_control.json}, \path{analysis/iclr/rebuttal_sf_scoring_sensitivity.json}, \path{analysis/iclr/persona_level_contrasts.json}, and the \path{traces/packaging_*} files.

\begin{table}[htbp]
\caption{Matched packaging control on the 520-turn core (GPT-5.1 / Sonnet 4.6). Overall counts all-turn passes under the judge; SF$_\text{key}$ counts SF passes of 360 under the answer key alone. The last columns give the 1k-to-10k change in the SF \ragmode-minus-\docmode gap under SF$_\text{key}$, against the \docmode rows of Table~\ref{tab:replay}, with persona-bootstrap 95\% intervals. The last row is the \ragmode row of Table~\ref{tab:replay} (model-written query, 120 answer tokens, original prompt), whose overall counts are not comparable.}
\label{tab:packaging}
\centering
\small
\setlength{\tabcolsep}{2.5pt}
\begin{tabular}{lrrrr}
\toprule
Excerpt & Overall & SF$_\text{key}$ & Gap change, GPT & Gap change, Sonnet \\
\midrule
$3\times450$ & 460 / 453 & 274 / 276 & $-24.7$ [$-48.7$, $-3.1$] & $-20.4$ [$-41.4$, $-0.1$] \\
$3\times550$ & 461 / 459 & 279 / 283 & $-24.7$ [$-47.4$, $-4.7$] & $-19.0$ [$-38.7$, $+0.2$] \\
Whole & 480 / 470 & 323 / 312 & $-6.7$ [$-22.7$, $+4.7$] & $-7.0$ [$-30.2$, $+14.1$] \\
1,650 total & 461 / 454 & 284 / 284 & $-25.4$ [$-49.3$, $-4.1$] & $-20.0$ [$-39.3$, $-1.2$] \\
\midrule
Table~\ref{tab:replay} & & 302 / 304 & $-24.0$ [$-48.3$, $-2.0$] & $-18.0$ [$-36.0$, $-2.0$] \\
\bottomrule
\end{tabular}
\end{table}

\paragraph{Open-source rows under the answer key.} The reruns of Table~\ref{tab:replay} differ from the submitted open-source rows in the embedder, the runner, and the SF prompt. \docmode uses no retrieval, yet judged \docmode SF passes rose between the two versions for Llama (314 to 334) and Gemma (296 to 315), while SF$_\text{key}$ passes did not (324 to 320; 310 to 310). Table~\ref{tab:os-key} therefore compares the versions under SF$_\text{key}$: the embedder narrows the SF loss by 5.0 to 6.1 points, and \ragmode still lowers SF accuracy for Llama and Gemma. Source artifact: \path{analysis/iclr/open_source_rerun_keyscored.json}; traces \path{t3_*} and \path{submitted_*}.

\begin{table}[htbp]
\caption{Open-source SF accuracy under SF$_\text{key}$ on the 360 SF turns: passes in the submitted rows and the reruns, and the \ragmode-minus-\docmode gap (points) with persona-bootstrap 95\% intervals.}
\label{tab:os-key}
\centering
\small
\setlength{\tabcolsep}{3pt}
\begin{tabular}{lrrrrr}
\toprule
Reader & \docmode & \ragmode & Gap, submitted & Gap, rerun & Change \\
\midrule
Qwen & 167 / 155 & 131 / 137 & $-10.0$ [$-20.5$, $+0.5$] & $-5.0$ [$-14.6$, $+4.7$] & $+5.0$ [$-6.1$, $+16.4$] \\
Llama & 324 / 320 & 246 / 264 & $-21.7$ [$-32.0$, $-11.8$] & $-15.6$ [$-25.8$, $-6.3$] & $+6.1$ [$+1.4$, $+11.6$] \\
Gemma & 310 / 310 & 248 / 269 & $-17.2$ [$-30.6$, $-4.3$] & $-11.4$ [$-22.7$, $-0.3$] & $+5.8$ [$+0.8$, $+11.9$] \\
\bottomrule
\end{tabular}
\end{table}

\subsection{Sonnet-written panel}
\label{app:sonnet-panel}

Sonnet 4.6 wrote the dossiers and the frozen items of a second replay panel under the same item schema. Its 16 personas are the first four per length bucket in a seeded order (seed 20260423) of the 35 Sonnet dossiers; the panel was planned at 8 personas and extended to 16 after the pilot. The panel yields 260 rated turns per condition (181 SF, 54 IF, 25 NIP). Both frontier readers ran \docmode, the \ragmode setting of Table~\ref{tab:replay} (model-written query, $3\times550$ characters), and \ragmode with whole sections; the two \ragmode runs can select different sections. No GPT model scores this panel: SF turns use the answer key alone, IF and judged NIP turns a Sonnet 4.6 judge, and other NIP turns refusal checks. The Sonnet judge grades compound SF answers with the IF rubric and accepts incomplete ones, so SF uses the key.

\begin{table}[htbp]
\caption{Sonnet-written panel (16 personas). Passes per condition and the 1k-to-10k change in the \ragmode-minus-\docmode gap (points) with persona-bootstrap 95\% intervals. The lower rows drop SF items flagged by the item check (179 turns remain).}
\label{tab:sonnet-panel}
\centering
\small
\setlength{\tabcolsep}{3pt}
\begin{tabular}{lrrrrr}
\toprule
& & \multicolumn{2}{c}{\ragmode} & \multicolumn{2}{c}{Gap change} \\
\cmidrule(lr){3-4}\cmidrule(lr){5-6}
Reader & \docmode & 550 & Whole & 550 & Whole \\
\midrule
GPT-5.1 & 216 & 192 & 214 & $-50.5$ [$-67.8$, $-21.2$] & $-6.2$ [$-13.9$, $+1.5$] \\
Sonnet 4.6 & 198 & 178 & 203 & $-35.1$ [$-66.9$, $+20.2$] & $+7.2$ [$-18.2$, $+43.8$] \\
\midrule
\multicolumn{6}{@{}l}{Flagged SF items removed (179 turns)} \\
GPT-5.1 & 175 & 146 & 161 & $-50.6$ [$-63.6$, $-17.6$] & $-6.1$ [$-15.6$, $+3.4$] \\
Sonnet 4.6 & 164 & 132 & 150 & $-50.6$ [$-62.1$, $-25.0$] & $-8.9$ [$-20.5$, $+3.1$] \\
\bottomrule
\end{tabular}
\end{table}

Persona-level contrasts (persona-bootstrap intervals and sign-flip tests over 16 personas): GPT-5.1 minus Sonnet as reader is $+6.9$ points under \docmode ($[2.7,13.5]$, $p=0.004$; 9 personas favor GPT and none favor Sonnet), $+5.4$ with excerpts ($p=0.004$), and $+4.2$ with whole sections ($p=0.12$). \ragmode with excerpts minus \docmode is $-9.2$ ($p=0.24$) and $-7.7$ ($p=0.33$); whole sections minus excerpts is $+8.5$ ($p=0.15$) and $+9.6$ ($p=0.03$). At 10k, retrieval selects a labeled section on 84.6\% (GPT) and 90.4\% (Sonnet) of turns, while the keyed SF answer is in 2.9\% and 0.0\% of excerpt packets and 62.9\% and 71.4\% of whole-section packets. Source artifacts: \path{analysis/iclr/sonnet_panel_analysis_nongpt.json}, \path{sonnet_panel_analysis_sfvalid_nongpt.json}, \path{persona_level_contrasts.json}, and the \path{traces/sonnet_panel_*} files.

\subsection{Item check}
\label{app:item-check}

Two validators, GPT-5.1 (minimal reasoning) and Sonnet 4.6 (temperature 0), read the full dossier, an SF question, and its keyed answer, without any reader output; an item is valid when both return ``supported''. On the main panel they flag 19 of the 93 rated SF items (66 of 360 SF turns); on the Sonnet-written panel, 16 of 44 (81 of 181). A typical flagged item is keyed to a fact from the persona's generation card that the dossier omits or states differently. Under \docmode, flagged main-panel turns pass 80.3\% for both readers, against 97.3\% (GPT) and 94.6\% (Sonnet) for the rest. Every reported value keeps all items; Table~\ref{tab:length-interaction} and Table~\ref{tab:sonnet-panel} give the results without flagged items. Source artifacts: \path{analysis/iclr/sf_item_validity_screen.json} and \path{sf_item_validity_table3.json}; trace field \texttt{sf\_item\_valid}.

\FloatBarrier
\end{document}